\documentclass[11pt]{article}

\usepackage[preprint]{acl}

\usepackage{times}
\usepackage{latexsym}
\usepackage{amsmath}
\usepackage{multirow}
\usepackage{pifont}
\usepackage{subfig}
\usepackage{multirow}
\usepackage{booktabs}

\usepackage[T1]{fontenc}
\usepackage[utf8]{inputenc}

\usepackage{microtype}

\usepackage{inconsolata}

\usepackage{graphicx}

\title{MisSpans: Fine-Grained False Span Identification \\ in Cross-Domain Fake News}

\author{%
  Zhiwei Liu\textsuperscript{1}\quad
  Paul Thompson\textsuperscript{1}\thanks{These authors contributed equally to this work.}\quad
  Jiaqi Rong\textsuperscript{2}\footnotemark[1]\quad 
  Baojie Qu\textsuperscript{3,4}\quad 
  Runteng Guo\textsuperscript{3,4}\quad \\
  \textbf{Min Peng}\textsuperscript{3,4}\quad
  \textbf{Qianqian Xie}\textsuperscript{3,4}\thanks{Corresponding author.}\quad 
  \textbf{Sophia Ananiadou}\textsuperscript{1,3,4,5} \\ 
    \textsuperscript{1}The University of Manchester \quad 
    \textsuperscript{2}Zhejiang University \quad \\
    \textsuperscript{3}School of Artificial Intelligence, Wuhan University \quad  \\
    \textsuperscript{4}Center for Language and Information Research, Wuhan University \quad 
    \textsuperscript{5}ELLIS Manchester \quad  \\
\texttt{\{zhiwei.liu,paul.thompson,sophia.ananiadou\}@manchester.ac.uk} \\
\texttt{xieq@whu.edu.cn}
}

\begin{document}
\maketitle
\begin{abstract}
Online misinformation is increasingly pervasive, yet most existing benchmarks and methods evaluate veracity at the level of whole claims or paragraphs using coarse binary labels, obscuring how true and false details often co-exist within single sentences. These simplifications also limit interpretability: global explanations cannot identify which specific segments are misleading or differentiate how a detail is false (e.g., distorted vs. fabricated).
To address these gaps, we introduce MisSpans, the first multi-domain, human-annotated benchmark for span-level misinformation detection and analysis, consisting of paired real and fake news stories.
MisSpans defines three complementary tasks: MisSpansIdentity for pinpointing false spans within sentences, MisSpansType for categorising false spans by misinformation type, and MisSpansExplanation for providing rationales grounded in identified spans. Together, these tasks enable fine-grained localisation, nuanced characterisation beyond true/false and actionable explanations.
Expert annotators were guided by standardised guidelines and consistency checks, leading to high inter-annotator agreement.
We evaluate 15 representative LLMs, including reasoning-enhanced and non-reasoning variants, under zero-shot and one-shot settings. Results reveal the challenging nature of fine-grained misinformation identification and analysis, and highlight the need for a deeper understanding of how performance may be influenced by multiple interacting factors, including model size and reasoning capabilities, along with domain-specific textual features.
%We introduce MisSpans, the first multi-domain benchmark with three complementary tasks and human-annotated high quality datasets for span-level misinformation detection. Firstly, false spans are identified within sentences, through comparison with real news stories. 
%Secondly, these spans are classified according to their misinformation type, and finally, an explanation of why each span constitutes misinformation is provided. MisSpans aims to facilitate more accurate, interpretable and transparent misinformation detection across domains. We have evaluated multiple open and closed source large language models, including both reasoning-based and no-think models, on all three tasks. Our results demonstrate the challenging nature of fine-grained misinformation identification and analysis, and highlight the need for a deeper understanding of how performance may be influenced by multiple interacting factors, including model size and reasoning capabilities, along with domain-specific textual features. 
\end{abstract}

\section{Introduction}

\begin{figure}[t]
\centering
  \includegraphics[width=0.6\columnwidth]{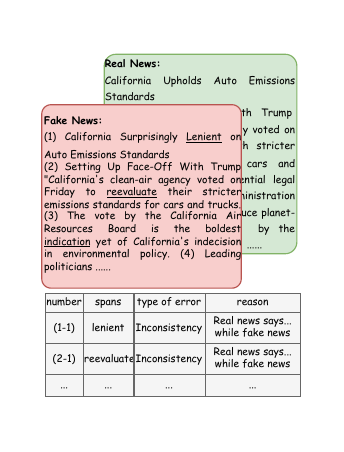}
  \caption{Example from the MisSpans. In each fake news sentence, false spans are identified, through comparison with the real news story, and assigned a misinformation type label. Finally, a reason why the span is considered to represent misinformation is provided.}
  \label{fig:mainmethod}
\end{figure}

With the rapid expansion of social networks, the online environment has become increasingly saturated with misinformation \cite{hilberts2025impact}. Unlike traditional fabricated stories that are entirely false, modern misinformation is often subtle and complex, interweaving accurate details with misleading or distorted elements \cite{chadwick2022deception}. This hybrid nature makes misinformation more persuasive and harder to detect for both humans and automated systems, especially since individual sentences can often include a mixture of genuine and false information.  Fine-grained misinformation analysis at the span level has the potential to address this challenge, by pinpointing the exact segments in which misinformation occurs and revealing which details are exaggerated, distorted or taken out of context \cite{wang2024factcheck}. 
%This can help to reduce the risks of misclassifying misinformation that is inherent when applying coarser-grained analyses. 
Span-level analysis can also greatly enhance interpretability, thus enabling fact-checking systems to provide clearer, evidence-based explanations \cite{zhang2025towards}. As such, automated fine-grained misinformation analysis represents a crucial step towards the construction of accurate, trustworthy and transparent fact-checking systems.

%There exist many benchmark datasets related to misinformation detection, covering tasks such as fact-checking \cite{kamoi2023wice,glockner2024ambifc}, fake news detection \cite{shu2020fakenewsnet,perez2018automatic}, rumour detection \cite{kochkina2018all,ma2017detect} and conspiracy detection \cite{langguth2023coco,miani2022loco}.

There exist many benchmark datasets related to misinformation detection, covering various domains, as shown in Table \ref{tab:relatedwork}. Despite their effectiveness, existing approaches suffer from three critical limitations. Firstly, they typically treat each claim or document as an atomic unit of analysis \cite{su2020motivations,zhou2020survey,liu2025raemollm}. This fails to capture the hybrid nature of modern misinformation, in which true and false information often co-exist within the same text. Secondly, the coarse-grained category labels used in most existing work (e.g., “true” or “false”) cannot distinguish between different types of false details, e.g., genuine information that has been distorted or manipulated vs. novel, fabricated information. Thirdly, explanations are usually provided at the global level; while useful for high-level assessment, they overlook the complex and subtle ways in which misinformation is expressed in text and do not reveal which specific segments are misleading \cite{fact2019report}. Many current approaches thus offer limited interpretability and actionable insight. Appendix \ref{app:relatedwork} contains further details of related work.

To address these shortcomings of existing datasets, we have develop \textit{MisSpans}, the first cross-domain benchmark with human-annotated datasets, for span-level misinformation detection and analysis, which consists of pairs of real and fake news stories with comparable content curated from authoritative news sources across multiple domains. Unlike existing benchmarks (see Table \ref{tab:relatedwork}), which primarily assess entire claims or paragraphs and provide global explanations, MisSpans introduces three complementary tasks with high-quality annotated supporting datasets aimed at directly tackling these limitations. Span identification (\textbf{MisSpansIdentity}) precisely pinpoints which spans within sentences constitute false information, thus supporting the accurate localisation of misleading content that is lacking in existing coarse-grained methods. Span type classification (\textbf{MisSpansType}) categorises the misinformation type represented by each false span, to enable more nuanced reasoning that moves beyond binary true/false labels. Finally, span explanation (\textbf{MisSpansExplanation}) provides rationales for why each span is misleading, to enhance interpretability and actionable insight. Data annotations relating to all three tasks were created by experts with strong linguistic expertise, who followed strict, standardised guidelines to ensure consistency, accuracy and high levels of inter-annotator agreement. 
%The annotation process involved firstly identifying  all false spans within each fake news story (MisSpansIdentity), secondly classifying each false span according to its misinformation type (MisSpansType), and finally authoring an explanation for each span (MisSpansExplanation), to aid in capturing the nuanced ways in which misinformation is expressed in text. 
\textit{MisSpans} thus constitutes a high-quality benchmark aimed at advancing both the accuracy and transparency of automated misinformation detection systems.

% We have evaluated the ability of a range of open and closed-source large language models (LLMs) to perform these three tasks, including models with enhanced reasoning capabilities. Our results show that current mainstream LLMs face substantial difficulties in identifying fine-grained misinformation spans and their types, thus demonstrating the challenging nature of the tasks, and the need for further research into more accurate automated methods.  

We evaluate 15 representative mainstream large language models (LLMs) on MisSpans, including both preliminary and reasoning-enhanced variants. Our results reveal several key insights. Firstly, all models struggle to accurately identify fine-grained misinformation spans, highlighting the inherent difficulty of this task. Secondly, reasoning-enhanced models generally outperform their non-reasoning counterparts of comparable size on complex tasks, although the benefits of reasoning capabilities are inconsistent across both tasks and domains. Thirdly, while one-shot prompting can slightly improve performance for some large-scale models, it often has limited or even negative effects for other models. This emphasizes the need for strategies that involve a greater degree of domain adaptation than one-shot prompting can allow. These findings collectively demonstrate the challenging nature of fine-grained misinformation identification and analysis, and point to the need for a deeper investigation into how performance may be influenced by multiple interacting factors. These include model size and reasoning capabilities, along with domain-specific features of news stories. The outcomes of such an investigation will help to drive the development of more accurate automated approaches to our novel tasks. 

Our main contributions are as follows:

\begin{itemize}
    
 \item We introduce MisSpans, the first multi-domain benchmark with human-annotated datasets for fine-grained span-level detection and analysis in fake news.
 %which also classifies the type of misinformation and provides corresponding explanations of why the spans constitute false information.

 \item We provide a comprehensive evaluation of the abilities and limitations of a wide range of LLMs to perform each task. 
 %The results show that current LLMs struggle to accurately identify fine-grained misinformation spans.

 \item We analyse the performance differences between zero-shot and one-shot settings and examine model reasoning errors, thus providing valuable insights to guide future research.
 % Our results show that current LLMs perform suboptimally on fine-grained false span detection, highlighting the need for more advanced methods for practical applications.

\end{itemize}

%To this end, we have developed MisSpans, the first multi-domain, human-annotated textual news dataset aimed at supporting the automated identification and analysis of precise textual spans that represent misinformation. It is intended that MisSpans will provide significant scope to enhance both the interpretability and accuracy of misinformation detection models. 

\section{MisSpans Benchmark}

\begin{figure}[t]
\centering
  \includegraphics[width=\columnwidth]{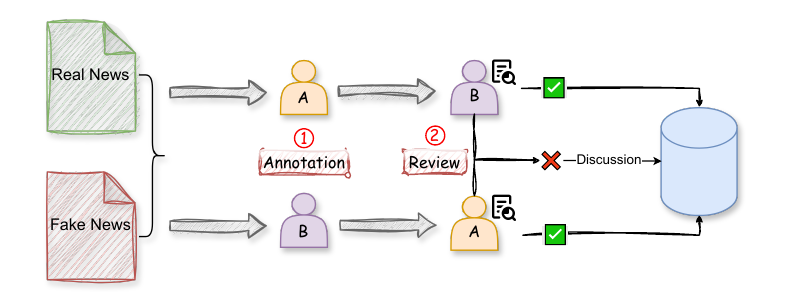}
  \caption{Overview of the process for labelling the MisSpans dataset. \ding{172} Annotation stage: each annotator labels half the dataset. \ding{173} Review stage: Each annotator labels the other half of the dataset and compares with the other annotator's labels to create a consolidated set.}
  \label{fig:mainmethod}
\end{figure}

%MisSpans provides a robust evaluation benchmark for assessing LLMs on fine-grained misinformation span detection, misinformation type classification, and associated explanations. 
This section outlines the MisSpans construction pipeline (see Figure \ref{fig:mainmethod}) and provides a detailed description of the task definitions, data construction process and quality validation procedures.

\subsection{Task Formulation}

Within the context of LLMs, the formal starting point for each data item in MisSpans is a real news article $P_{real}$ and a fake news article $P_{fake}$, where $P_{fake}$ consists of n sentences $\{s_1, s_2,...s_n\}$. Each sentence $s_i$ contains $m$ words $\{w_{i,1}, w_{i,2}, w_{i,m}\}$. 

\textbf{Task 1. MisSpansIdentity.} The aim is to identify all spans $\{w_{i,begin}-w_{i,end}\}$ that convey misinformation in each sentence of the fake news article, based on comparison with the information presented in the real news article.

\begin{equation}
\small
Spans = \arg\max_{Spans} P_{\mathrm{LLM}}(Spans|P_{real},P_{fake})
\end{equation}

\textbf{Task 2. MisSpansType.} The aim is to assign a misinformation category to each identified span.

\begin{equation}
\small
\mathcal{C} = \{\text{True},\ \text{Inconsistency},\ \text{No mention},\ \text{Others}\},
\end{equation}

\begin{equation}
\small
\hat{c} = \arg\max_{c \in \mathcal{C}} P_{\mathrm{LLM}}(C|P_{real},P_{fake},Spans).
\end{equation}

\textbf{Task 3. MisSpansExplanation.} The aim is to provide a corresponding explanation of why each identified span is considered to convey misinformation. Each item in $E$ is $e_{i,k}$.

\begin{equation}
\small
E = \arg\max_{E} P_{\mathrm{LLM}}(E|P_{real},P_{fake},Spans,C)
\end{equation}

\subsection{Data Collection}

The basis for developing MisSpans is the FakeNewsAMT dataset \cite{perez2018automatic}, consisting of real news stories paired with fake news counterparts. This dataset was chosen according to its potential to support robust misinformation detection methods; the multiple domains covered provide evidence of how the topics covered, and hence, the potential types of misinformation spans, can vary across different domains. For example, fake business stories may report company profits rather than losses, while false technology stories may exaggerate or invent features of newly released products, etc. 

FakeNewsAMT consists of two subsets. The Crowdsourced subset contains 40 news stories in each of six different domains (sports, business, entertainment, politics, technology, and education), collected from a variety of mainstream news websites, including ABC News, CNN, USA Today and New York Times. 
%To ensure the quality and reliability of these stories,  the developers of FakeNewsAMT conducted manual fact-checking, by verifying the news source and cross-referencing information among several sources. T
For each real news story, the Amazon Mechanical Turk (AMT) platform was used to obtain a corresponding human-authored, fake story, with a similar writing style, topic and length to the real story. 

%For each real news article, AMT Workers were asked to produce both a fake headline and a fake news body, which imitated the news writing style of each story as closely as possible, and which covered the same topic and had a similar length to the original news item.  The complete Crowdsourced subset thus contains a total of 240 real news stories, each paired with a fake version generated by AMT workers.

The Web subset of consists of 100 story pairs about public figures 
%(actors, singers, socialites, and politicians) 
from entertainment publications such as Entertainment Weekly, People, and RadarOnline. 
%The data were collected in  true-false pairs. 
Rather than using AMT for false story generation, the Web part collected "naturally occurring" fake news. i.e., stories containing false information already circulating on the web. Stories were classified as real or fake using gossip fact-checking sites like GossipCop.com and through cross-checking with other entertainment sources. We selected a random subset of 40 story pairs for inclusion in MisSpans.
%In contrast to the Crowdsourced part, in which the fake news stories had been especially created by the AMT workers who followed specific instructions, the Web part aimed to collect "naturally occurring" fake news. i.e., stories containing false information that are were already circulating on the web. For the Web subset, the classification of each article as real or fake was determined through evaluation of the claims made using gossip fact-checking sites like GossipCop.com and through cross-checking with other entertainment sources. A total of 100 legitimate and 100 fake articles were collected. Compared to the Crowdsourced subset of the dataset, the Web articles tended to be much longer, with greater differences between the styles and content of the real and fake articles. According to these differences, considerably more time and effort were required for annotation that for the Crowdsouced subset. Due to the availability of limited human resources for our annotation effort, we thus decided to annotate a random set of 40 articles from the Web subset of FakeNewsAMT for inclusion in our MisSpans dataset. We refer to these articles as the \textit{celebrity} domain. 

\subsection{Task Design}

%The data collection process described above resulted in 280 real-fake news story pairs. 
Each of the 280 collected fake news stories was split into individual sentences (a total of 1,749 sentences). Annotators then focus on identifying misinformation on a sentence by sentence basis, through comparison to the information presented in the corresponding real news story, while still considering the contextual information during the process. We defined three separate annotation tasks, as described below. More detailed guidelines for these tasks can be found in Appendix \ref{app:guidelines}.

\subsubsection{Task 1: MisSpansIdentity}

For each sentence in the fake news story, annotators were asked to identify all spans conveying information that diverges from information in the corresponding real story, thereby indicating misinformation. The real story served as the sole reference for determining factual accuracy; external world knowledge was not to be used. For each piece of misinformation identified, the shortest possible text span within the fake story that clearly represents the misinformation was selected. 
%In cases where the misinformation corresponds to an inconsistency between the information presented in the real and fake articles, the selected span should precisely capture the detail that differs between two articles. 
If all information in a particular sentence was deemed to be completely consistent with the information presented in the real story, then the label ``True'' was assigned to the complete sentence, in place of any specific spans.   

\subsubsection{Task 2: MisSpansType}

For each span identified in Task 1, a category label is chosen, according to the type of misinformation that the span represents (if any). We defined the following four possible labels:   

\begin{itemize}
    
\item \textbf{True}: If the sentence was labeled  as “True,” for Task 1, then the label in this task should also be “True.”
\item \textbf{Inconsistency}: Assigned when both the real and fake content refer to the same topic, event, or entity (e.g., a person, object, organization, or location), and the misinformation span corresponds to details about the topic, event or entity that differ between the fake and real content. Since cases of inconsistencies between real and fake content can be very varied, the annotation guidelines include a range of indicative examples. Further details are provided in Appendix \ref{app:inconsistency}.
\item \textbf{No mention}: Assigned when the misinformation span in the fake sentence cannot be aligned with any information the real content. 
%If the sentence contains multiple “no mention” parts, it should not be further divided; a single label is sufficient.
\item \textbf{Others}: Assigned when a fake content span is indicative of misinformation, but none of the above labels are applicable. Examples include grammatical or spelling errors. 
\end{itemize}
Two of these labels are intended to identify different techniques that fake news authors may use to deliberately use deceive readers, i.e., altering, manipulating or distorting details present in the original news story (\textit{Inconsistency}) and fabricating new information (\textit{No mention}). Spans in the \textit{Other} category usually constitute ``red flags'' that a news story may originate from an unreliable source.

\subsubsection{Task 3: MisSpansExplanation}

A brief explanation is provided to justify why each identified span is considered to constitute misinformation. For spans assigned the  \textit{Inconsistency} or \textit{No Mention} labels in Task 2, annotators followed templates when formulating the reasons, in an attempt to enhance consistency and interpretability. Further details can be found in Appendix \ref{app:guidelines}.

%The templates are as follows: 

%\textbf{For Inconsistency:}
%Real content says “[real content]”, [which expresses … (optional)], while fake content says “[fake content]”, [which expresses … (optional)].

%\textbf{For No mention:}
%“There is no mention of [fake content] in the real content.”

%If multiple “no mention” parts occur within one sentence, they should be separated and described individually.

\subsection{Annotation}

% The annotation was carried out by two annotators with experience in linguistics.
The annotation work was performed by two professionals, each with extensive expertise in syntax, semantics and discourse analysis, which were considered important to perform the tasks accurately. Further details about their backgrounds are provided in Appendix \ref{app:annotaorexpertise}. 
%in an attempt to ensure a high level of linguistic accuracy and consistency.  
Following the development of initial annotation scheme and associated guidelines, an iterative, multi-round preliminary annotation phase aimed to refine and improve the scheme and guidelines, so as to maximise annotation quality and consistency in the final dataset. Each round of this phase used a common set of 30 real-fake news story pairs covering all seven domains, and proceeded as follows:
%(four each from the six domains in the Crowdsourced subset, and six from the Celebrity domain). Using these samples, we followed an iterative process to improve the quality of both the annotations and the guidelines, as follows: 
\begin{itemize}
    \item Each annotator independently created or modified their annotations based on the current version of guidelines.
    \item In a cross-review process, each annotator compared their own annotations with those of the other annotator to create a final, consolidated set of annotations for each news story pair.
    \item Each annotator's consolidated version was compared with both their original annotations and the consolidated version of the other annotator as the basis for a discussion about discrepancies and their reasons for occurring. 
    %For example, if one annotator originally missed an annotation that was identified by the other annotator, was this a simple oversight or due to a lack of clarify on the guidelines? Also, if discrepancies occurred between the two annotators' consolidated versions, how could the disagreement be resolved?
    \item Following the discussion, guidelines were modified, e.g., to include clearer explanations, rules and/or additional illustrative examples, aimed at improving ease and consistency of annotator decisions in the next round.
\end{itemize}

When agreement between the two annotators on the 30 sample stories had stabilised, the final dataset was produced as follows:
\begin{itemize}
    \item \textbf{Independent annotation:} The remaining 250 stories were split into two halves, and each annotator independently labelled one half. 
    \item \textbf{Cross-validation:} For each item in the other half of the dataset, each annotator firstly performed independent labelling, and then immediately compared their own annotations with those of the other annotator to create a final, consolidated set of annotations. Although time and labour constraints meant that each story could only be cross-validated by a single annotator, this process ensured that cross-validation was performed while each story was fresh in the annotators' minds.  Despite substantial improvement in the agreement between annotators during the course of the preliminary phase, this stage was still considered important to ensure the highest possible quality of the final dataset. 
    %Given the diversity of topics and styles covered by the article in the dataset, combined with the high density of individual pieces of misinformation in many of the articles, annotation was sometimes very complex and labour-intensive, and it was impossible or the guidelines to cover all potentially difficult cases. As such, it was expected that annotator may still miss certain annotations that the other annotator identified, or that the two annotators would sometimes make divergent annotation decisions. 
    Any disagreements were collectively discussed to reach a final decision.
\end{itemize}

%An example of the potential complexity of annotating a single sentence in the fake article, consider the following sentences from a real and corresponding fake article in the \textit{Education} domain of the Crowdsourced subset cound be found in Appendix \ref{app:annotationexample}.

Appendix \ref{app:annotationexample} shows an example of the potential complexity of annotating a fake story sentence.

\subsubsection{Data Quality Validation \label{sec:dataquality}}

\begin{table}[]
\footnotesize
\resizebox{0.5\textwidth}{!}{
\begin{tabular}{p{0.8cm}lcccc}
\hline
             & Task          & Acc   & F1    & Kappa & Similarity \\ \hline
Prelimi-nary  & Task 1\_Exact & 0.897 & 0.755 & -     & -          \\
             & Task 1\_Relax & 0.949 & 0.854 & -     & -          \\
             & Task 2        & 0.921 & 0.955 & 0.866 & -          \\
             & Task 3        & -     & -     & -     & 0.811      \\ \hline
Random Items & Task 1\_Exact & 0.907 & 0.703 & -     & -          \\
             & Task 1\_Relax & 0.966 & 0.846 & -     & -          \\
             & Task 2        & 0.957 & 0.959 & 0.925 & -          \\
             & Task 3        & -     & -     & -     & 0.924      \\ \hline
\end{tabular}
}
\caption{Inter-annotator agreement scores for each task.}
  \label{tab:agreementscore}
\end{table}

Table \ref{tab:agreementscore} reports the inter-annotator agreement scores at the end of the preliminary annotation phase, and from a randomly selected subset of 30 stories from the final annotation effort. 
Although good levels of agreement were reached by the end of the the initial annotation phase, annotators still sometime struggled to achieve consistent decisions about whether to choose longer ``spanning'' \textit{No Mention} annotations or more fine-grained ``split'' \textit{Inconsistency} annotations (as exemplified in Appendix \ref{app:disagreements}); either decision can often be argued to be correct. Therefore, we randomly selected 30 stories from the final set, and focused specifically on cross-validating cases in which there was a ``spanning'' vs ``split'' discrepancy. If both annotators' decisions were considered valid, we deemed the split version to be correct. The ``Random Items'' section of Table \ref{tab:agreementscore} reports agreement following the resolution of this challenging type of discrepancy.
For Task 1, inspired by evaluation methods in named entity recognition \cite{wang2021nero}, we designed two evaluation strategies: \textbf{Exact span matching} indicates the extent to which both annotators select exactly the same misinformation spans, while \textbf{Relaxed span matching} requires only that spans chosen by the two annotators include some overlapping part\footnote{Detailed calculation examples are provided in Appendix~\ref{app:evaluation_spans}.}.  Relaxed span matching helps to quantify anotators' agreement on individual pieces of misinformation, even if their exact spans differ. Although the annotation guidelines include rules about the content of annotated spans (e.g., that prepositions should be excluded from the start or end of spans), the wide range of types of spans that can denote misinformation, along with different phrasing or sentence structures between real and fake stories, can sometimes result in difficult span-level decisions. 

Similarly to named entity annotation, span selection tasks do not include negative examples. Therefore, for Task 1, we report the pairwise F1 score as an alternative to Cohen’s Kappa coefficient, since this is commonly used to report agreement in named entity tasks. Task 2 is a classification task, and we report the F1 and Kappa scores. For Task 3, we measure inter-annotator agreement of the explanations using sentence similarity (all-mpnet-base-v2 \cite{song2020mpnet}). 

The results in Table \ref{tab:agreementscore} show that, although exact misinformation span selection is clearly a challenging task, relaxed span agreement is much higher. This indicates that the annotators mostly agree on identifying individual pieces of misinformation, if not their exact spans. 
Agreement for both Tasks 2 and 3 is high, and the statistics provide strong evidence that the generally good levels of agreement attained by the end of the preliminary annotation phase were maintained or even improved in the final annotation phase. 

%Discrepancies often concern disagreements between the \textit{Inconsistency} and \textit{No mention} categories, particularly when the structure of the fake news story diverges from that of the real story, and/or where multiple entities, events or topics could potentially form the focus of an inconsistency. An example of such a discrepancy is provided and discussed in detail in Appendix \ref{app:disagreements} 

% \subsection{Dataset Analysis}

% Some statistic analysis on multidomain

\section{Experimental Setup}

\subsection{Experimental LLMs}

Our extensive evaluation of the ability of open-source and proprietary LLMs to perform the three tasks included the following reasoning-focused models: gpt-5 \cite{openai_gpt5_system_card_2025}, DeepSeek-V3.2-Reasoner \cite{deepseek_models}, Claude-Sonnet-4.5 \cite{claude_opus45}, Gemini-2.5-Flash \cite{gemini_api_docs}, and Qwen3 reasoning variants (8B-R, 14B-R, and 32B-R) \cite{qwen3_techreport}. For comparison, we also assessed a diverse set of no-think LLMs, i.e., GPT-4.1 \cite{openai_gpt41}, DeepSeek-V3.2-Chat \cite{deepseek_models}, the Qwen3 no-think models (8B, 14B, and 32B) \cite{qwen3_techreport}, Qwen2.5-72B-Instruct (Qwen72B) \cite{qwen25_official}, Llama-3.1-8B-Instruct, and Llama-3.3-70B-Instruct \cite{llama31_techreport}.

\subsection{Experimental settings and metrics}

\textbf{Settings:} We evaluated the LLMs using two prompting settings. In the zero-shot setting, the LLMs were directly provided with the task description and requirements. In the one-shot setting, a single example from the same domain as the story to be analysed served as a reference from which the LLMs could learn. Detailed prompt templates can be found in Appendix \ref{app:prompttemplate}. Due to the complexity of the tasks, they were evaluated in a step-by-step manner: Task 2 was tested using the gold-standard outputs of Task 1, and Task 3 was tested using the gold-standard outputs of both Tasks 1 and 2\footnote{Since Tasks 2 and 3 build upon Task 1, and the LLMs did not perform satisfactorily on Task 1, we do not report results where the LLMs were required to complete all three tasks simultaneously.}.

\textbf{Metrics:} For Task 1, we show the exact and relaxed F1 scores, as described in Section \ref{sec:dataquality}. Task 2 is a standard classification task, for which we report Accuracy and F1 scores. For Task 3, we used ROUGE-L \cite{lin2004rouge} and the sentence similarity metric described in Section \ref{sec:dataquality}.

\section{Experimental Results}

\begin{table*}[]
\resizebox{1\textwidth}{!}{
\begin{tabular}{lccccccccccccccccc}
\hline
                  & \multicolumn{8}{c}{\textit{\textbf{Zero-shot}}}                                                                             & \multicolumn{1}{l}{\textit{\textbf{}}} & \multicolumn{8}{c}{\textit{\textbf{1-shot}}}                                                                                \\ \cline{2-9} \cline{11-18} 
Models            & \multicolumn{2}{c}{Task 1}      &           & \multicolumn{2}{c}{Task 2}      &           & \multicolumn{2}{c}{Task 3}      &                                        & \multicolumn{2}{c}{Task 1}      &           & \multicolumn{2}{c}{Task 2}      &           & \multicolumn{2}{c}{Task 3}      \\ \cline{2-3} \cline{5-6} \cline{8-9} \cline{11-12} \cline{14-15} \cline{17-18} 
                  & Exact          & Relax          &           & Acc            & F1             &           & Rouge          & Sim            &                                        & Exact          & Relax          &           & Acc            & F1             &           & Rouge          & Sim            \\ \hline
gpt-5             & 0.267          & 0.371          &           & 0.547          & 0.366          &           & 0.618          & 0.748          &                                        & 0.269          & 0.374          &           & 0.599          & 0.410          &           & 0.633          & 0.789          \\
DeepSeek-Reasoner & \textbf{0.302} & \textbf{0.429} & \textbf{} & 0.593          & 0.381          &           & 0.619          & 0.751          &                                        & \textbf{0.304} & \textbf{0.433} & \textbf{} & 0.629          & 0.434          &           & 0.634          & 0.779          \\
claude-sonnet-4.5 & 0.300          & \textbf{0.429} & \textbf{} & \textbf{0.644} & \textbf{0.455} & \textbf{} & 0.638          & \textbf{0.796} & \textbf{}                              & 0.294          & 0.413          &           & \textbf{0.649} & \textbf{0.480} & \textbf{} & 0.640          & \textbf{0.802} \\
gemini-2.5-flash  & 0.287          & 0.420          &           & 0.462          & 0.307          &           & 0.593          & 0.729          &                                        & 0.280          & 0.400          &           & 0.493          & 0.316          &           & 0.610          & 0.766          \\
Qwen3-8b-R        & 0.279          & 0.388          &           & 0.510          & 0.320          &           & 0.528          & 0.649          &                                        & 0.284          & 0.398          &           & 0.486          & 0.312          &           & 0.499          & 0.614          \\
Qwen3-14b-R       & 0.271          & 0.375          &           & 0.543          & 0.351          &           & 0.573          & 0.690          &                                        & 0.256          & 0.338          &           & 0.528          & 0.334          &           & 0.561          & 0.684          \\
Qwen3-32b-R       & 0.271          & 0.379          &           & 0.528          & 0.345          &           & 0.546          & 0.660          &                                        & 0.269          & 0.373          &           & 0.529          & 0.343          &           & 0.533          & 0.660          \\
gpt-4.1           & 0.291          & 0.426          &           & 0.593          & 0.389          &           & 0.614          & 0.768          &                                        & 0.289          & 0.419          &           & 0.617          & 0.412          &           & 0.634          & 0.796          \\
DeepSeek-Chat     & 0.286          & 0.398          &           & 0.628          & 0.436          &           & \textbf{0.649} & 0.764          &                                        & 0.296          & 0.409          &           & 0.645          & 0.474          &           & \textbf{0.641} & 0.776          \\
Qwen3-8b          & 0.211          & 0.285          &           & 0.598          & 0.407          &           & 0.601          & 0.748          &                                        & 0.216          & 0.294          &           & 0.544          & 0.343          &           & 0.506          & 0.639          \\
Qwen3-14b         & 0.270          & 0.381          &           & 0.520          & 0.345          &           & 0.603          & 0.733          &                                        & 0.273          & 0.385          &           & 0.440          & 0.322          &           & 0.609          & 0.746          \\
Qwen3-32b         & 0.251          & 0.356          &           & 0.548          & 0.351          &           & 0.583          & 0.717          &                                        & 0.262          & 0.364          &           & 0.582          & 0.350          &           & 0.598          & 0.745          \\
Qwen2.5-72b       & 0.268          & 0.377          &           & 0.595          & 0.375          &           & 0.642          & 0.785          &                                        & 0.271          & 0.384          &           & 0.598          & 0.404          &           & 0.638          & 0.792          \\
LLama8b           & 0.212          & 0.314          &           & 0.549          & 0.377          &           & 0.481          & 0.678          &                                        & 0.223          & 0.310          &           & 0.323          & 0.279          &           & 0.362          & 0.516          \\
LLama70b          & 0.281          & 0.391          &           & 0.562          & 0.357          &           & 0.598          & 0.740          &                                        & 0.243          & 0.317          &           & 0.593          & 0.396          &           & 0.609          & 0.769          \\ \hline
\end{tabular}
}
\caption{Overall performance on three tasks. ``sim" denotes the  similarity score.\label{tab:overall_3tasks}}
\end{table*}

\begin{figure*}[t]
\centering
  \includegraphics[width=1.6\columnwidth]{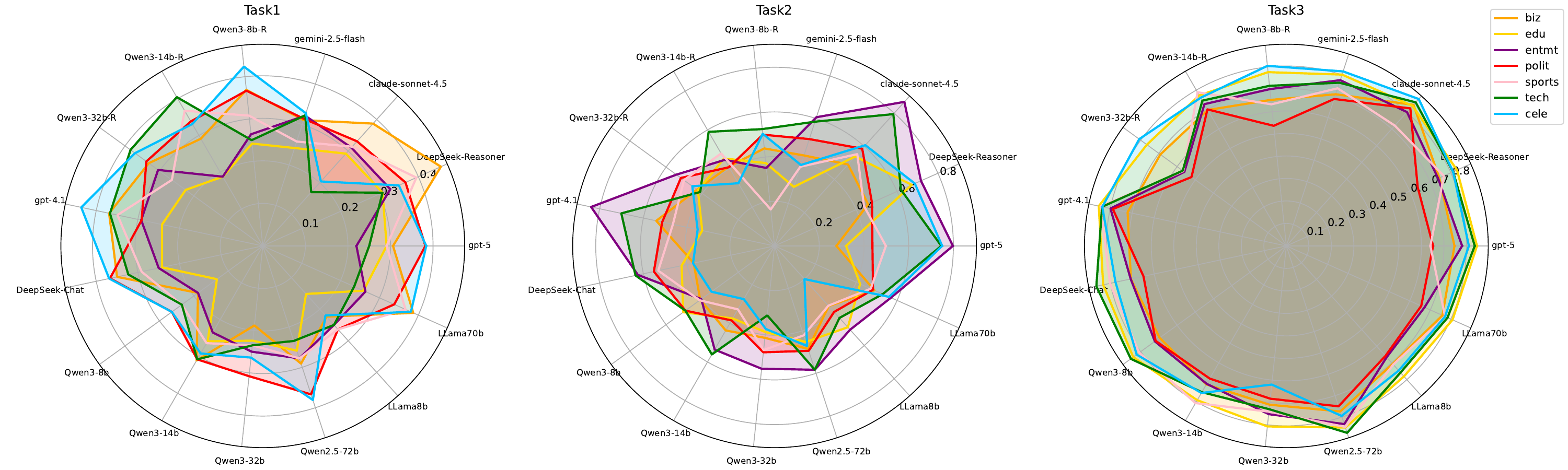}
  \caption{Radar charts comparing model performance across domains in zero-shot setting. Task 1: exact score, Task 2: F1 score. Task 3: similarity score. The complete results can be found in Tables \ref{tab:task1_domsins}, \ref{tab:task2_domsins},\ref{tab:task3_domsins}.}
  \label{fig:radar_zeroshot}
\end{figure*}

% \begin{figure*}[t]
% \centering
%   \includegraphics[width=2\columnwidth]{figs/oneshot_figure.pdf}
%   \caption{Radar charts comparing model performance across domains in one-shot setting. Task 1: exact score, task 2: F1 score. Task 3: similarity score.}
%   \label{fig:radar_zeroshot}
% \end{figure*}

\subsection{Main Findings}

Table \ref{tab:overall_3tasks} presents the overall performance of each LLM on the three tasks. The MisSpansIdentity and MisSpansType tasks are demonstrated to be highly challenging, with F1 scores for all models falling below 0.5. In contrast, as expected, LLMs perform relatively well on the MisSpansExplanation task. Although models with enhanced reasoning capabilities (e.g., DeepSeek-Reasoner and Claude-Sonnet-4.5) demonstrate the strongest and most robust general performance, such capabilities do not seem uniformly advantageous across all tasks and domains. A further finding is that one-shot prompting should be used with caution, give its potential to be harmful to all but the largest and/or most advanced models, compared to zero-shot prompting. While there are significant cross-domain performance differences for the MisSpansIdentity and MisSpansType tasks (see Figure \ref{fig:radar_zeroshot}), these do not always correlate with model size or reasoning abilities. This suggests that multiple factors may influence optimal performance in certain domains. 

\subsection{Task 1: MisSpansIdentity}

\textbf{All evaluated models struggle to accurately identify misinformation spans correctly}. Misinformation spans can be very diverse and high performance in this task relies on acquiring a thorough understanding of the content of both the real and fake stories, followed by a detailed analysis of each sentence in the fake article, to determine whether and how potentially multiple pieces of misinformation are expressed in the sentence.
%and which text spans should be used to represent them.

The majority of spans identified by the best performing models appear to contain misinformation, suggesting that recall is more problematic than precision. It seems particularly challenging for models to identify several separate pieces of misinformation that occur within a single sentence, particularly when these belong to multiple misinformation types, such as the example shown in \ref{app:multispanexample}. 
%An examination of the output of Deepseek-Reasoner shows that, although it initially identifies two \textit{Inconsistency} spans and two \textit{No Mention} spans as it analyses the fake news sentence in a stepwise manner, it untimately seems to ``forget'' the \textit{No mention spans}, and ultimately outputs only the \textit{Inconsistency} spans. 
Subtle differences are sometimes completely overlooked, e.g., one fake article refers to ``Pizzagate" as a \textit{controversy}, which softens its description as \textit{hoax} in the corresponding real article. 
%There can also be a misconception that in the case of inconsistencies, misinformation spans can correspond to the spans that are common between the real and fake versions of an article (e.g., specific entities), rather than spans that identify the inconsistency in the description of the common entity. 
Other spans frequently missed by the models include typographic and grammatical errors, especially when these do not occur in proper names (e.g., people and places).

For all models, performance improves when using the relaxed evaluation criterion. 
%In general, many misinformation spans identified by the models tend to be longer than the gold standard ones. 
For example, while some gold standard spans consist of single nouns, verbs and adjectives, the corresponding model-identified spans often consist of complete noun phrases or verbs and their objects, even if some words are shared between the real and fake articles. An example is shown in Appendix \ref{app:task1modeloutput}.

\textbf{For models of comparable size (e.g., Qwen3 and DeepSeek series), reasoning variants generally seem more adept than their no-think counterparts.} A notable exception is GPT-5, whose lower performance may be due to its specialisation for code-related tasks. Deepseek-Reasoner appears particularly proficient in identifying fine-grained spans, while other models are more often prone to identifying longer spans that should be split into several shorter spans. More explicit guidelines in prompts may help to better harness the power of reasoning models. For example, the DeepSeek-Reasoner output shows that certain misinformation spans (e.g., spelling errors) are often noted, but are not counted as misinformation spans.   

\textbf{One-shot prompting leads to slight improvements for only a small subset of models (e.g., DeepSeek and Qwen no-think series), and leads to performance degradation for most models}. Given the wide variety of lengths and structures of misinformation spans, which can range from single words up to complete sentences, providing only a small number of examples in one-shot prompts appears to confuse, rather than help most models. For example, the output of Claude-Sonnet-4.5 in the zero-shot setting sometimes results in finer-grained misinformation spans than the one-shot setting.

\textbf{Pronounced performance differences exist between some individual domains, especially in the zero-shot setting, and for certain domains, larger model size does not always lead to better performance (see Table \ref{tab:task1_domsins} and Figure \ref{fig:radar_zeroshot}).} In the zero-shot setting, the best performances using the exact metric in the business (DeepSeek-Reasoner), technology (Qwen3-14B-R), and celebrity (Qwen3-8B-R) domains are relatively strong (above 0.4). The top relaxed metric results in the politics (DeepSeek-Reasoner) and sports (gemini-2.5-flash) domains exceeding 0.6. While DeepSeek-Reasoner achieves comparatively strong performance across most domains, the large performance differences between the politics (0.664) and technology domains (0.350) may be due to differences in the localisation difficulty of misinformation spans, with sentences in technology stories tending to have a higher number of misinformation spans than those in the political domain.
%reflecting variations in the linguistic structure and localization difficulty of misinformation spans. For example, sentences in technology stories tend to have a higher number of misinformation spans than those in the political domain.  Notably, larger model size does not always lead to better performance in certain domains, such as technology, celebrity, and sports.

\subsection{Task 2: MisSpansType}

\textbf{Performance differences are weakly correlated with model size, but more strongly related to model alignment strategies and training objectives}. For example, while Claude Sonnet 4.5 and DeepSeek-Chat achieve relatively strong performance,
%followed by Qwen-8B
other large-scale LLMs do not exhibit particularly strong results; notably, Gemini-2.5-Flash had the lowest performance. For such multi-class classification tasks, models optimised for reasoning stability and instruction-following capabilities (e.g., Claude Sonnet and DeepSeek) outperform those primarily designed for low latency and general-purpose interaction (e.g., Gemini Flash\footnote{https://storage.googleapis.com/deepmind-media/Model-Cards/Gemini-2-5-Flash-Model-Card.pdf}). 
% \textcolor{red}{(Need to double-check the source.)}

\textbf{For models of comparable size, reasoning-oriented models (with the exception of Claude Sonnet 4.5) tend to underperform their no-think counterparts}. A possible reason is that reasoning-oriented models may ``overthink'' the label assignment process, or apply criteria that are too strict. For example, the \textit{Inconsistency} label will often apply in cases where there are sentences with comparable structures in the real and fake stories, but where one or more of the structural elements (e.g., subject, verb or object) has been changed. While such surface-level differences appear to be straightforward for many models to classify as \textit{Inconsistency}, DeepSeek-Reasoner sometimes tends to focus too much on ``fine details'', e.g. an inconsistency in the subject or object of a comparable sentence will only be labelled as such if the entity types match between the real and fake stories. A specific example is provided in Appendix \ref{app:task2modeloutput}

\textbf{One-shot prompting only appears to be beneficial for large-scale LLMs}. For example, a small number of examples appears sufficient for DeepSeek-Reasoner to noticeably refine its approach (+5.3\% vs. zero-shot). In contrast, the performance decline of smaller LLMs in the one-shot setting (e.g., the Qwen3 series and LLaMA-8B) suggests that they lack the ability to reliably abstract task rules from a limited number of examples, and instead appear to be distracted by the specific content of the examples.

\textbf{As with Task 1, there can be significant performance differences between different domains, although the higher scores indicate that misinformation classification is generally an easier task (see Table \ref{tab:task2_domsins} and and Figure \ref{fig:radar_zeroshot})}. For the zero-shot results, the entertainment (claude-sonnet-4.5), technology (claude-sonnet-4.5) and celebrity (gpt-5) domains achieve particularly high F1 scores, possibly due to their pattern-driven misinformation characteristics. e.g., many fake technology stories tend to retain similar sentence structures to the real stories, but alter details such as prices or features of products.  Overall, Claude-Sonnet-4.5 demonstrates strong cross-domain performance, suggesting superior robustness in holistic semantic judgment for misinformation classification.

\subsection{Task 3: MisSpansExplanation}

%Results are analysed using similarity scores.
%\textbf{All models perform better than the first two tasks, especially the large-scale LLMs}. This indicates that current generative models are particularly suited to  generation tasks such as this. 

%Several of the models appear to e very competent in identifying text spans in real articles that are inconsstent with those in fake articles. For example, given the misinformation span  \textit{will be challenged} and the category \texitt{Inconsistency}, the Deepseek-Reasoner output is \textit{Real content says "brands are keen to see a true rival emerge to challenge Facebook and Google", while fake content says "Facebook and google will be challenged."}. However, part of the task is to explain \textit{why} the span in the fake news constituties. In this respect, Claude Claude Sonnet 4.5, even in its zero-shot setting, often does a better job, and in this case outputs \textit{Real content says "Brands are also keen to see a true rival emerge to challenge Facebook and Google", which describes brands' desire for a challenge, while fake content says "Facebook and google will be challenged", which presents it as a certainty}. Such examples help to explain why Claude Sonnet 4.5 achieves such high scores for this task.  
\textbf{Similarly to Task 2, reasoning models of comparable size (e.g., DeepSeek and Qwen3 series) achieve lower performance than their no-think counterparts}. This may be because the reasoning models carry out complex and sometimes unnecessary reasoning steps, which have the potential to introduce errors, e.g., trying to determine \textit{why} a misinformation span has been categorised as \textit{No Mention}, rather than simply generating an appropriate explanation using the template provided. In contrast, the reliance of no-think models on direct input-to-output mappings often results in more stable explanations. Nevertheless, there is evidence that reasoning abilities can help to generate more descriptive and helpful explanations of inconsistencies, as exemplified in Appendix \ref{app:task3modeloutput}.
%While the inconsistency explanations for some models simply identify the text span in the real article that is inconsistent with the misinformation span, the Claude Sonnet 4.5 output often explains \textit{why} there is an inconsistency, e.g., \textit{Real content says "Brands are also keen to see a true rival emerge to challenge Facebook and Google", which describes brands' desire for a challenge, while fake content says "Facebook and google will be challenged", which presents it as a certainty}.

%This suggests that, for the misinformation explanation task, explicit reasoning may introduce intermediate noise that smaller models cannot reliably handle, whereas no-think models rely on direct input-to-output mappings, resulting in more stable explanations.

\textbf{With the exception of Qwen-8B (-R) and LLaMA-8, one-shot prompting improves performance over to zero-shot prompting}. Similarly to Task 2, these results indicate that large-scale models can make effective use of a limited number of examples to improve task alignment and generate more stable explanations. For example, while Claude Sonnet 4.5 is able to generate some very useful descriptive explanations even in the zero-shot setting, such descriptions become more numerous and often exhibit increased clarify in the one-shot setting. In contrast, however, smaller models may be distracted by the additional context, leading to performance degradation.

\textbf{Performance across different domains is more stable for most models compared to other tasks (see Table \ref{tab:task3_domsins})}.  The best zero-shot similarity scores for all models in each domain exceed 0.8, thus providing strong evidence that current mainstream LLMs already possess mature and stable cross-domain capabilities for explanation generation.

\section{Conclusion}

This paper has introduced MisSpans, the first cross-domain, human-annotated benchmark aimed at supporting the detection, classification and explanation of span-level misinformation. We have provided a detailed description of these three tasks and the annotations created to support them. Our evaluation of 15 mainstream LLMs using MisSpans indicates that the tasks of identifying and classifying misinformation spans represent significant challenges, even for advanced closed-source LLMs. A detailed analysis of the results has revealed strengths and limitations of current LLMs when applied to each task. Collectively, our novel MisSpans benchmark, experimental results and insights gained constitute a valuable basis to drive and guide future research into more accurate and transparent
misinformation detection across different domains.

\section*{Limitations}

Due to computational resource limitations, we have only evaluated open-source LLMs with sizes up to approximately 70B parameters. 
%Therefore, we did not consider the performance of larger model architectures on this benchmark. 
For similar reasons, we did not fine-tune LLMs on the MisSpans dataset. However, we believe that domain-specific fine-tuning would likely improve their performance. 

Due to the chosen base dataset of real-fake pairs, the current data is available only in English and is not multilingual. This may restrict its applicability to other languages.

% \section*{Acknowledgments}

%  \emph{International Joint Conference on Artificial Intelligence} and the \emph{Conference on Computer Vision and Pattern Recognition}.


\bibliography{main}

\appendix

\section{Related Work \label{app:relatedwork}}

\subsection{Benchmarks for misinformation detection}

Benchmark datasets for misinformation detection cover a range of different tasks. Datasets for fact-checking are typically used to develop methods to verify the truthfulness of information, either by using a retriever to search a large corpus (e.g., \cite{wadden2022scifact,ma2024ex, wang2024factcheck}) for the most relevant evidence to support claim verification or correction, or by having a model verify claims based on the provided documents \cite{kamoi2023wice,glockner2024ambifc}. Fake news detection typically involves analysing long-form text, often collected from sources such as PolitiFact\footnote{https://www.politifact.com/} and GossipCop\footnote{https://www.gossipcop.com/}, or using datasets such as Fakenewsnet \cite{shu2020fakenewsnet}, FakeNewsAMT \cite{perez2018automatic}, and KaggleNews \cite{megan_risdal_2016}. Rumour detection datasets are usually based on data from social media platforms, containing the source post, comments and the corresponding social network structure; examples include PHEME \cite{kochkina2018all} and Twitter15/16 \cite{ma2017detect}. Conspiracy theory detection typically involves using datasets such as \citet{langguth2023coco,miani2022loco} to automatically identify whether a text contains narratives about ``secret forces controlling events or concealing the truth.”.  There are additionally many misinformation-related datasets aimed at instruction tuning LLMs \cite{han2025exploring,liu2024conspemollm,liu2025fmdllama,Rangapur2023FinFactAB}. 

Despite this plethora of existing misinformation datasets, they generally focus on the overall truthfulness of the complete data items (e.g., articles or posts). They neglect to account for the fact that, in many cases only part(s) of the complete information conveyed in the data item is actually false.

\subsection{Benchmarks for span detection}

There exist many benchmark datasets to support the development of methods to detect different types of text spans.  Examples include question-answering datasets (e.g., \cite{rajpurkar2016squad, trischler2017newsqa} which identify text spans within articles that represent answers to specific questions. The MPQA dataset \cite{wiebe2005annotating} annotates subjectivity and opinion spans in news commentary and is used for opinion extraction and sentiment analysis. The CoNLL 2012 \cite{pradhan2012conll} multilingual corpus annotates entity spans, coreference chains and syntactic structures, to support research into named entity recognition and coreference resolution. Further datasets to support named entity recognition tasks include NCBI Disease \cite{dougan2014ncbi}, CLUENER \cite{xu2020cluener2020}, and WNUT-17 \cite{derczynski2017results}. 

Datasets annotated with fine-grained spans could also play an important part in developing more accurate and useful misinformation detection methods. Matos \cite{de2024towards} aimed to address this by developing datasets in which the exact parts of  Brazilian Portuguese videos containing false claims were identified. However, to our knowledge, no similar datasets have been developed for textual sources, in which the information conveyed in individual sentences may include both false and genuine details. 

\begin{table*}[htb]
\resizebox{\textwidth}{!}{
\begin{tabular}{llcccl}
\hline
Task                 & Datasets                              & Spans Identity & Classification & Explanation & Level              \\ \hline
Fact-check           & Scifact \cite{wadden2022scifact}      & \ding{55}     & \ding{51} & \ding{55}   & Sentence           \\
Fact-check           & EX-FEVER \cite{ma2024ex}              & \ding{55}     & \ding{51} & \ding{51}   & Sentence           \\
Fact-check           & Factcheck \cite{wang2024factcheck}    & \ding{55}     & \ding{51} & \ding{55}   & Sentence, document \\
Fact-check           & WICE \cite{kamoi2023wice}             & \ding{55}     & \ding{51} & \ding{55}   & Sentence, document \\
Fact-check           & AmbiFC \cite{glockner2024ambifc}      & \ding{55}     & \ding{51} & \ding{55}   & Sentence           \\
Fact-check           & FinFact \cite{Rangapur2023FinFactAB}  & \ding{55}     & \ding{51} & \ding{55}   & Sentence           \\
Fake News Detection  & Fakenewsnet \cite{shu2020fakenewsnet} & \ding{55}     & \ding{51} & \ding{55}   & document           \\
Fake News Detection  & FakeNewsAMT \cite{perez2018automatic} & \ding{55}     & \ding{51} & \ding{55}   & document           \\
Rumor Detection      & PHEME \cite{kochkina2018all}          & \ding{55}     & \ding{51} & \ding{55}   & Sentence           \\
Rumor Detection      & Twitter15/16 \cite{ma2017detect}      & \ding{55}     & \ding{51} & \ding{55}   & Sentence           \\
Conspiracy Detection & COCO \cite{langguth2023coco}          & \ding{55}     & \ding{51} & \ding{55}   & Sentence           \\
Fact-check           & MisSpans (ours)                       & \ding{51}     & \ding{51} & \ding{51}   & word-level         \\ \hline
\end{tabular}
}
\caption{Comparison of misinformation datasets by task category, text granularity and support for span identification, classification and explanation. \ding{51} indicates support, \ding{55} indicates lack of support. \label{tab:relatedwork}}
\end{table*}

\section{Annotation Details}

\subsection{Guidelines \label{app:guidelines}}

\begin{center}
\footnotesize
\fcolorbox{black}{gray!10}{
\begin{minipage}{0.45\textwidth}
\footnotesize
\textbf{Guidelines for Task 1: MisSpansIdentity}  \\

\textbf{Identify misinformation spans in fake news articles:}

Identify any information in the fake news that differs from the real article, indicating misinformation. The real article should be used as the sole source of accurate information to identify misinformation in the fake news. External world knowledge should not be used.

\textbf{Labeling sentences for misinformation:}

Each numbered fake news sentence should have at least one associated label. Each piece of misinformation should be numbered as (n1-n2), where n1 is the sentence number and n2 is the misinformation count in the sentence (e.g., (1-1), (1-2)). If the fake news sentence aligns completely with information in the real article, label it as (n1-1),  e.f.g. (2-1), followed by the word 'True'.

\textbf{Misinformation span selection:}

Select the shortest text span in the fake news that clearly identifies misinformation. If the misinformation concerns an inconsistency, the span should identify the detail that differs between the real and fake content. 

\end{minipage}
}
\end{center}

\begin{center}
\footnotesize
\fcolorbox{black}{gray!10}{
\begin{minipage}{0.45\textwidth}
\footnotesize
\textbf{Guidelines for Task 2: MisSpansType}  \\

\textbf{Format:} [\textit{number}] [\textit{Misinformation category}]

(1) The number is the same as the number marked in task 1.

(2) Misinformation category: True, Inconsistency, No mention, Others

\textbf{True:} If the annotation is “True” in Task 1, the label here is also True.

\textbf{Inconsistency:} This category should be assigned when both the real and fake content mention a common topic, event or entity (e.g., a person, object/thing, company, or location), but one or more of the details regarding the common topic, entity, or event are different in the fake content, compared to the real content.  

\textbf{No mention:} This category should be used only if no parts of the fake sentence align with information in the real content. If the sentence contains multiple “no mention” parts, do not need to split it into many numbers, just use one number.

\textbf{Others:} If none of the above labels are suitable, use the ‘Others’ label.

\end{minipage}
}
\end{center}

\begin{center}
\footnotesize
\fcolorbox{black}{gray!10}{
\begin{minipage}{0.45\textwidth}
\footnotesize
\textbf{Guidelines for Task 3: MisSpansExplanation}  \\

A short explanation of why the specific category label was assigned to the piece of misinformation. 

\textbf{For inconsistency}, the reason should have the following format: 
Real content says “[real content]”, [which expresses…this is optional], while fake content says “[fake content]”,  [which expresses…this is optional].

\textbf{For No mention label}: The reason template should mention which parts of the fake content do not occur in the real content, e.g., "There is no mention of the whole selected sentence/specific information in real content". If multiple no-mention parts are in one sentence, split and describe them separately.

\end{minipage}
}
\end{center}

\subsection{Annotator Expertise \label{app:annotaorexpertise}}

The annotation was carried out by two experts with the following expertise: 

\textbf{Annotator 1}: A researcher with a computational linguistics background and considerable experience of research into different areas of natural language processing (NLP), including misinformation, with an emphasis on developing annotated datasets to support various NLP tasks across multiple domains.  

\textbf{Annotator 2}: An undergraduate student majoring in English Language and Literature. With a basic understanding of fake news and linguistics, this annotator contributes to the annotation of false spans in fake news from a research-oriented perspective.

\subsection{Inconsistency examples \label{app:inconsistency}}

Cases of inconsistencies between real and fake content can be very varied. In order to provide annotators with an overview of the potential scope of the types of differences between real and fake content that should be classified as inconsistencies, the guidelines list and exemplify a range of possible ways in which the various details about entities and events can vary between the real and fake content. The guidelines make it clear that these only represent illustrative examples, and that further types of inconsistencies may also be encountered. These illustrative examples are listed below and are exemplified in Table \ref{tab:examplesannotation}. 

\textbf{Example 1:} There is a difference in an action, event, or relationship in which one or more entities that are common to the real and fake content are involved.

\textbf{Example 2:}  There is a difference in the specification of when an event took place that is common to the real and fake content.

\textbf{Example 3:} There is a difference in one or more properties/characteristics of an event or entity that is common to the real and fake content.

\textbf{Example 4:} There is a difference in the entities that are involved in an event that is common to the real and fake content.

\textbf{Example 5:} A description/relationship relating to an entity or event that is common to the real and fake content has been exaggerated or toned down in the fake content.

\textbf{Example 6:} There is a difference in the polarity or certainty of an event that is common to the real and fake content (e.g., an event that is stated to have taken place in the real content is negated in the fake content, or vice versa).

% \begin{table}[]
% \footnotesize
% \resizebox{0.48\textwidth}{!}{
% \begin{tabular}{lll}
% \hline
% Source        & Entity                                            & Associated Event                                                    \\ \hline
% \textbf{Real} & {\color[HTML]{00B050} \textbf{Day Without Women}} & {\color[HTML]{FFC000} \textbf{Two districts cancel school}}         \\
% \textbf{Fake} & {\color[HTML]{00B050} \textbf{Day Without Women}} & {\color[HTML]{FFC000} \textbf{Male teachers prepare to cover down}} \\ \hline
% \end{tabular}
% }
% \end{table}

\begin{table*}[]
\resizebox{\textwidth}{!}{
\begin{tabular}{p{15cm}lp{4cm}p{5cm}}
\hline
Example 1                                                                                                                                                                                                     & Source & Entity                                            & Associated Event                                                               \\
On 'Day Without Women,' Two Districts Cancel School                                                                                                                                                           & Real   & Day Without Women                                 & Two districts cancel school                                                    \\
Male teachers preparing to cover down on 'Day Without Women'                                                                                                                                                  & Fake   & Day Without Women                                 & Male teachers prepare to cover down                                            \\ \hline
Example 2                                                                                                                                                                                                     & Source & Time                                              & Associated Event                                                               \\
Samsung   Galaxy Note 7 phones will be banned from all airline flights after nearly 100 incidents of the devices overheating and sometimes   injuring owners, the Transportation Department announced Friday. & Real   & Friday                                            & ban of devices on flights                                                      \\
On   Monday, the Federal Aviation Administration announced they were expanding the   ban to include all smart phones and tablets                                                                              & Fake   & Monday                                            & ban of devices on flights                                                      \\ \hline
Example 3                                                                                                                                                                                                     & Source & Entity                                            & Property                                                                       \\
Alex   Jones Apologizes for Promoting 'Pizzagate' Hoax                                                                                                                                                        & Real   & Pizzagate                                         & hoax                                                                           \\
Alex   Jones Vindicated in "Pizzagate" Controversy                                                                                                                                                            & Fake   & Pizzagate                                         & controversy                                                                    \\ \hline
Example 4                                                                                                                                                                                                     & Source & Entity 1                                          & Action                                                                         \\
Basketball   'bible' auction sets sports memorabilia record                                                                                                                                                   & Real   & Basketball 'bible'                                & sets                                                                           \\
Signed   autograph of Michael Jordan's 'Basketball' auction sets sports memorabilia   record                                                                                                                  & Fake   & Signed autograph of Michael Jordan's 'Basketball' & sets                                                                           \\ \hline
Example 5                                                                                                                                                                                                     & Source & Entity                                            & Relationship                                                                   \\
Alex Jones  a prominent conspiracy theorist                                                                                                                                                                   & Real   & Jones                                             & {[}is{]}                                                                       \\
Jones, who has been accused by many   mainstream media outlets of being a conspiracy theorist)                                                                                                                & Fake   & Jones                                             & accused of being                                                               \\ \hline
Example 6                                                                                                                                                                                                     & Source & Modification/negation                             & Event                                                                          \\
The Pizzagate theory  which posited with no evidence that top   Democratic officials were involved with a satanic child pornography ring                                                                      & Real   & no evidence                                       & top Democratic officials were involved with a satanic child   pornography ring \\
Jones and others uncovered evidence last   year that top Democratic Party officials were involved in a bizarre, satanic   child sex cult and pornography ring                                                 & Fake   & uncovered evidence                                & top Democratic officials were involved with a satanic child   pornography ring \\ \hline
\end{tabular}
}
\caption{Examples for annotation. \label{tab:examplesannotation}}
\end{table*}

\subsection{Annotation examples \label{app:annotationexample}}

\subsubsection{Example of a fake sentence with multiple misinformation spans \label{app:multispanexample}}

This example shows how even a fairly simple sentence in a fake news story can contain multiple misinformation spans of different types, thus helping to motivate our fine-grained approach to misinformation span detection and categorisation. In the following example, all relevant information to be compared with the information in the fake news sentence is contained within a single sentence of the real article. However, it should be noted that, in many cases, the annotation task is more complex, because the structure and sentence content of the fake story do not always mirror the real article so closely. Accordingly, the complete content of the real article must be considered to correctly determine whether any parts of a fake news sentence contain misinformation  

\textbf {Fake article sentence}: \textit{(3) Almost 70 percent of US colleges and universities have reported that there has been an increase by at least 1\% of the amount of international applications recieved in the past 2 years.}

\textbf{Real article sentence}: \textit{Nearly 40 percent of responding U.S. institutions are reporting a drop in international student applications, particularly from students in the Middle East, according to initial findings from a survey of 250 schools.}

\textbf{Task 1 spans}

(3-1) 70

(3-2) US colleges and universities

(3-3) increase

(3-4) at least 1

(3-5) recieved

(3-6) in the past 2 years

\textbf{Task 2 labels and Task 3 reasons}

(3-1) \textit{Inconsistency}: Real content says that findings are based on reports from nearly 40 percent of responding U.S. institutions, while fake content says that findings are based on reports from almost 70 percent of of US colleges and universities

(3-2) \textit{Inconsistency}: Real content says that the survey concerns 250 schools, while fake content says that it concerns all US colleges and universities

(3-3) \textit{Inconsistency}: Real content says that U.S. institutions are reporting a drop in international student applications, while fake content says that there has been an increase in applications

(3-4) \textit{No mention}: There is no mention of the percentage change in the number of applications in real content

(3-5) \textit{Other}: The word “received” in fake content is misspelled as “recieved”

(3-6) \textit{No mention}: There is no mention that the reports are based on applications received in the past two years in real content

The fake sentence contains example of all three types of misinformation spans, as follows: 

\textbf{Inconsistency}

Both the real and fake news items report information about a common entity, i.e., a statistic regarding findings about the change in the number of international student applications received in US educational establishments  However, there are three different instances of inconsistencies regarding the exact details of this statistic between the two versions of the news story, i.e. cases where comparable, but incompatible, details are specified about the statistic: 
\begin{itemize}
\item Firstly, (3-1) identifies that there is a numerical difference in the reported statistic between the real and fake content, While the real content states that the reported finding concerns nearly 40 \% of institutions, the fake content states that the finding concerns almost 70 \% of institutions. 
\item Secondly, (3-2) establishes that there is a difference in the scope of the findings between the two versions of the story - the statistic in the real news story is based on responses from a survey of only 250 schools, while the fake content gives the impression that the statistic is based on all US colleges and universities. 
\item  Thirdly, (3-3) reveals that there is a difference in the directionality of the change reported in the real and fake items - while the real news story reports a drop in international student applications, the fake news reports an increase in applications. 
\end{itemize}

\textbf{No Mention}

In contrast to the inconsistencies identified by annotations (3-1), (3-2) and (3-3),  annotations (3-4) and (3-6) identify details about the statistic that are only specified in the fake news item, i.e. there are no comparable details in the real news story, thus resulting in their categorisation as \textit{No Mention}.  
\begin{itemize}
\item Annotation (3-4) identifies that in the fake news, the magnitude of the reported change in the number of applications (i.e. at least 1\%) is reported, but no such change in magnitude occurs in the real content. Note that, prior to assigning the \textit{No Mention} category, annotators should ensure that the detail does not appear elsewhere in the real article, even when there are sentences in the real and fake stories that have comparable content and/or structure, as the case for the current example. Although we do not show the rest of the real article here, there no mention of a change in the magnitude of change anywhere in the real story. 
\item Annotation (3-6) establishes that further detail that is missing from the real content is the time span over which the change in the number of applications is reported to have taken place (i.e., the past two years).  Again, such a time span does not occur anywhere within the real article.
\end{itemize}

\textbf{Other}

Annotation (3-6) identifies a further word from the sentence, which does not fit into the other two misinformation categories, namely a spelling error of the word “received” as “received”. Since news stories from genuine sources are likely to be carefully proofread, articles which contain spelling or grammatical errors, particularly in multiple places, can act as further flags that the news story is not genuine. 
% Apart from the 30 pre-annotated samples, each annotator labeled 250 of the remaining 250 samples and reviewed 125 of them. 

\subsubsection{Annotation disagreements \label{app:disagreements}}

 It can sometimes be particularly challenging to distinguish between information in the fake article that cannot be aligned at all with information in the real article (and hence should be assigned the \textit{No Mention} label), and information that corresponds to a deliberate manipulation/distortion of information presented in the real article (and hence should be be assigned the \textit{Inconsistency} label). Either choice can often be argued to be correct, since the information presented in sentences can sometimes be ``viewed' in different ways. However, the specific choices made by by the annotator may lead to disagreement in terms of the number and lengths of misinformation spans identified. In this section, we provide and discuss an example of such a case that led to disagreements about whether to create single, longer ``spanning'' annotations or multiple shorter ``split'' annotations.  

 \begin{center}
\footnotesize
\fcolorbox{black}{gray!10}{
\begin{minipage}{0.45\textwidth}
\footnotesize
\textbf{Real Story}  \\

Bruno: Convicted of murdering ex-girlfriend

Goalkeeper makes controversial return to soccer. After serving seven years in prison for killing his ex-girlfriend and feeding her to dogs  Brazilian goalkeeper Bruno Fernandes de Souza is controversially back in the game  signed by Boa Esporte for two years. Fans and sponsors of the Brazilian second-division side quickly denounced the move but so far Boa Esporte isn't backing down. In a lengthy post on its Facebook page  Boa Esporte's president  Rone Moraes da Costa says the team isn't committing any crimes by signing the 32-year-old who formerly played for one of Brazil's most famous clubs -- Flamengo -- and was tipped to line up for the national team at the 2014 World Cup on home soil.

\end{minipage}
}
\end{center}

\begin{center}
\footnotesize
\fcolorbox{black}{gray!10}{
\begin{minipage}{0.45\textwidth}
\footnotesize
\textbf{Fake Story}  \\

Bruno Fernandes de Souza Exonerated at Last

In a triumph of justice over misguided and insular ideas of human rights, popular soccer hero Bruno de Souza has at last been freed from prison. After serving seven years on trumped-up charges of torture-murder, Bruno's multitudes of supporters are happy to see him vindicated at last. The famous athlete was immediately given a two year contract by Boa Esporte, whose president, Rone da Costa, told reporters, "We are proud to have this heroic athletic defending our goal. He should never have been taken off the field. Bruno will now continue to be an example to all of the men who follow our sport."

\end{minipage}
}
\end{center}

The example concerns an article in the sports domain about Brazilian goalkeeper Bruno Fernandes de Souza. The real article explains that de Souza has been signed by the team Boa Esporte after serving seven years in prison for killing his ex-girlfriend, but that fans and sponsors of Boa Esporte have denounced this signing. 

The corresponding fake article begins with the headline \textit{Bruno Fernandes de Souza Exonerated at Last}, giving the false impression that he was not guilty of the crime, while a later sentence in the fake article describes him as a \textit{popular soccer hero}, which is inconsistent with the fans' denouncement of him. While both annotators agree that both of the above constitute inconsistencies with the real article, a disagreement ocures  regarding the following span in the fake article: \textit{Bruno's multitudes of supporters are happy to see him vindicated at last}. While one annotator chose the span \textit{multitudes of supporters are happy to see him vindicated at last}, and labelled it as \textit{No Mention}, the other annotator identified two separate spans, i.e., \textit{multitudes of supporters} and \textit{vindicated}, and labelled them both as \textit{Inconsistency}. 

The first annotator argued that de Souza's fans are not explicitly mentioned in the real article, which is indeed the case: only the fans of the team Boa Esporte are mentioned in the real article, as stated above. Therefore, strictly, the mention of de Souza's own fans and their attitudes/opinions cannot be directly aligned with information in the real article, and hence the ``spanning'' annotation \textit{multitudes of supporters are happy to see him vindicated at last} was identified and assigned the \textit{No Mention} label. 

The second annotator instead decided that \textit{multitudes of supporters} constitutes an inconsistency with the real article. It seems likely that the fake news author has chosen the phrase \textit{multitudes of supporters} to directly contradict the mention of de Souza's denouncement by fans of Boa Esporte in the real article. Nevertheless, as mentioned above, it could be argued that \textit{fans of Boa Esporte} are a separate group of people from \textit{Bruno's multitudes of supporters} and so do not necessarily constitute a common entity between the two versions of the article. However, de Souza \textit{is} a common entity between both stories, and the way in which he is portrayed in the fake story (i.e., as a popular hero who was wrongly convicted) is completely inconsistent with his portrayal on the real article (i.e., as a convicted criminal who is deeply unpopular). Thus, when considering de Souza as the entity in focus, the mention of him having multitudes of supporters is clearly aimed falsely emphasising his popularity. Similarly, the span \textit{vindicated} falsely suggests that he was innocent of the convicted crime.  
Although there is a case for deeming either annotator's analyses as correct, since the information presented in the sentence may be ``viewed'' in different ways, the more fine-grained ``split'' choices of the second annotator appear better aligned with the overall goals of the task. This split analysis establishes links between information on the real and fake content, and thus helps to build evidence of numerous and potentially subtle range devices that fake news authors may use to manipulate the facts presented in real articles in an attempt to convince readers to believe in false information. 

It is also interesting to note that both inconsistencies identified by the second annotator were identified by different LLMs. For example, Claude Sonnet identified \textit{supporters are happy to see him vindicated} as a misinformation span. While this span is similar to the longer span identified by the first annotator, the reasoning provided by Claude Sonnet 4.5 suggests that it considers this to be an \textit{Inconsistency} rather than a \textit{No Mention}, i.e., \textit{The real article also states fans denounced the move, not that "multitudes of supporters are happy to see him vindicated."}. Thus, the contrast between de Souza's denoucement by fans and the happiness of his multitudes of supporters has been noted by the model. However, this model does not explicitly say anything about \textit{vindication}.  On the other hand, Deepseek-Reasoner identifies \textit{vindicated at last} as a misinformation span, but does not identify the span \textit{multitude of supporters}.

The fact that the two LLMs have, between them, identified two separate pieces of misinformation here also strengthens the evidence that the second annotator's ``split'' analysis is the more correct one. Nevertheless, the observation that neither LLM could identify \textit{both} pieces of misinformation underlines the challenges of identifying misinformation at such a fine-grained level.  

\section{Evaluation of Exact and Relaxed Matching in Task 1 \label{app:evaluation_spans}}

Consider the sentence $\{w_{i,1}, w_{i,2}, w_{i,3}, ... w_{i,m}\}$. Annotator A selects $\{w_{i,2}^A, w_{i,3}^A,w_{i,4}^A, w_{i,5}^A\}$, while Annotator B selects $\{w_{i,4}^B, w_{i,5}^B,w_{i,6}^B, w_{i,7}^B\}$. The Exact span matches will split them into three parts:

$\{w_{i,2} - w_{i,3}\}$: {A: 1, B: 0},

$\{w_{i,4} - w_{i,5}\}$: {A: 1, B: 1},

$\{w_{i,6} - w_{i,7}\}$: {A: 0, B: 1},

While a relaxed span match will consider the spans selected by the two annotators as consistent (i.e. A: 1, B: 1). 

One specific example is ``Vaccines cause serious long term side effects in children.'' Annotator A selects ``cause serious long term'', while Annotator B selects ``long term side effects''.
The exact span matching will split the annotations into three parts:

``cause serious'': {A: 1, B: 0},

``long term'': {A: 1, B: 1},

``side effects'': {A: 0, B: 1}.

While a relaxed span matching will consider the spans selected by the two annotators as consistent (i.e., A: 1, B: 1), as both annotations target the same misinformation claim with overlapping boundaries.

After collecting all resulting split spans, we compute the F1 score over these spans for evaluation.

\section{Prompt Templates \label{app:prompttemplate}}
% (\ref{fig:propmt_task1},\ref{fig:propmt_task2},\ref{fig:propmt_task3},\ref{fig:propmt_1shot})

\begin{figure*}[htb]
\centering
\footnotesize
\fcolorbox{black}{gray!10}{
\begin{minipage}{\textwidth}
\footnotesize

\textbf{Zero-shot prompt for Task 1} \\
\\
Task description: Identify misinformation spans in fake news articles:
You will receive: \\
1. A REAL news article (the only source of truth) \\
2. A FAKE news article with numbered sentences \\
\\
Your task is to identify any information in the fake news that differs from the real article, indicating misinformation. The real article should be used as the sole source of accurate information used to identify misinformation in the fake news. External world knowledge should not be used. \\
\\
Labeling sentences for misinformation: \\
Each numbered fake news sentence should have at least one associated label. Each piece of misinformation should be numbered as (n1-n2), where n1 is the sentence number and n2 is the misinformation count in the sentence (e.g., (1-1), (1-2)). If the fake news aligns completely with information in the real article, label it with the sentence number (n1-1) followed by the word 'True'. \\
\\
Misinformation span selection: \\
Select the shortest text span in the fake news that clearly identifies misinformation. If the misinformation concerns an inconsistency, the span should identify the detail that differs between the real and fake content. If a sentence contains multiple errors, please list them separately according to the numbering requirements. \\
\\
STRICT FORMAT REQUIREMENTS: \\
1. Output MUST strictly follow this structure: \\
    (1-1) <misinformation span or True> \\
    (1-2) <misinformation span if exists> \\
    (2-1) <misinformation span or True> \\
    (2-2) ...
    ...\\
2. Every line MUST begin with the pattern: (number-number) \\
3. No extra commentary, no explanation, no justification. \\
4. Do NOT add missing or invented numbers. Only output numbers that correspond to the fake news sentence numbers. \\
5. Each (n1-n2) MUST be placed on its own separate line. \\
6. You must not output anything outside the specified format. \\
\\
Real news: \\
\textit{[real news]} \\
Fake news: \\
\textit{[fake news]}

\end{minipage}
}
\label{fig:propmt_task1}
\end{figure*}

\begin{figure*}[htb]
\centering
\footnotesize
\fcolorbox{black}{gray!10}{
\begin{minipage}{\textwidth}
\footnotesize

\textbf{Zero-shot prompt for Task 2} \\
\\
Task description: Categorise the type of misinformation spans

Misinformation spans refer to the specific false segments identified in the fake news, annotated sentence by sentence using the real news as the sole reference for accuracy. \\
\\
Label category: True, Inconsistency, No mention, Other \\
- True: If the misinformation spans annotation is “True”, the label here is also True. \\
- Inconsistency: This category should be assigned when both the real and fake content mention a common topic, event, or entity (e.g., a person, object/thing, company, or location), but one or more of the details regarding the common topic, entity, or event are different in the fake content, compared to the real content. \\
- No mention: This category should be used only if no parts of the fake sentence align with information in the real content. \\
- Other: If none of the above labels are suitable, use the ‘Other’ label. \\
\\
STRICT FORMAT REQUIREMENTS: \\
1. Output MUST strictly follow the pattern: \\
(1-1) <category> \\
(1-2) <category> \\
(2-1) <category> \\
(2-2) ... \\
2. Each line must begin with the `(n1-n2)` format corresponding to the span numbers from misinformation spans. \\
3. No additional commentary, explanation, or text is allowed. \\
4. Do NOT add or remove any numbers. Only output numbers corresponding to misinformation spans numbers. \\
5. Each `(n1-n2)` must be on its own line. \\
6. Output ONLY the category labels for each numbered span. \\
\\
Real news: \\
\textit{[real news]} \\
Fake news: \\
\textit{[fake news]} \\

Misinformation spans: \\
\textit{[Gold spans]}
\end{minipage}
}
\label{fig:propmt_task2}
\end{figure*}

\begin{figure*}[htb]
\centering
\footnotesize
\fcolorbox{black}{gray!10}{
\begin{minipage}{\textwidth}
\footnotesize

\textbf{Zero-shot prompt for Task 3} \\
\\
Task description: Provide an explanation of why the specific category label was assigned to the piece of misinformation spans. \\

Misinformation spans refer to the specific false segments identified in the fake news, annotated sentence by sentence using the real news as the sole reference for accuracy. Misinformation category is the false type of misinformation spans. \\
\\
EXPLANATION TEMPLATES: \\

1. For the “Inconsistency” label: 
   The explanation MUST follow this template: \\
   Real content says “[real content]”, [which … optional], while fake content says “[fake content]”, [which … optional]. \\
2. For the “No mention” label:
   The explanation MUST explicitly mention absent information, using this template: \\
   There is no mention of [the whole selected sentence / specific fake information] in real content. \\
3. For the “Others” label: \\
   - Grammatical error:
     "..." in fake content is ungrammatical; it should be "..." \\
   - Spelling error:
     The word "..." is misspelled as "..." in fake content. \\
   - Other cases:
     Prefer the template: Real content says "...", while fake content says "..." \\
     If unsuitable, provide a clear and concise explanation. \\

STRICT FORMAT REQUIREMENTS: \\
1. You MUST output explanations strictly in the form: \\
   (1-1) <explanation> \\
   (1-2) <explanation> \\
   (2-1) <explanation> \\
   ... \\
2. Each line MUST begin with a parenthesised index identical to Misinformation spans and Misinformation category (n1-n2). \\
3. No extra commentary, no justification, no introduction, no summary. \\
4. Do NOT add or remove any indices. Use only the indices provided. \\
5. Each (n1-n2) explanation MUST be placed on a separate line. \\
6. Explanations MUST follow the templates exactly. \\
\\
Real news: \\
\textit{[real news]} \\
Fake news: \\
\textit{[fake news]} \\

Misinformation spans: \\
\textit{[Gold spans]} \\

Misinformation category: \\
\textit{[Gold labels]}

\end{minipage}
}
\label{fig:propmt_task3}
\end{figure*}

\begin{figure*}[htb]
\centering
\footnotesize
\fcolorbox{black}{gray!10}{
\begin{minipage}{\textwidth}
\footnotesize

\textbf{1-shot prompt for Task 1} \\
\\
\textit{[Zero-shot prompt for Task 1]} (same as Zero-shot prompt for Task 1)

Here is an annotation example: \\
Real news: \\
\textit{[real news]} \\
Fake news: \\
\textit{[fake news]} \\
Misinformation spans: \\
\textit{[Gold spans]} \\

\textbf{1-shot prompt for Task 2} \\
\\
\textit{[Zero-shot prompt for Task 2]} (same as Zero-shot prompt for Task 2)

Here is an annotation example: \\
Real news: \\
\textit{[real news]} \\
Fake news: \\
\textit{[fake news]} \\
Misinformation spans: \\
\textit{[Gold spans]} \\
Misinformation category: \\
\textit{[Gold labels]} \\

\textbf{1-shot prompt for Task 3} \\
\\
\textit{[Zero-shot prompt for Task 3]} (same as Zero-shot prompt for Task 3)

Here is an annotation example: \\
Real news: \\
\textit{[real news]} \\
Fake news: \\
\textit{[fake news]} \\
Misinformation spans: \\
\textit{[Gold spans]} \\
Misinformation category: \\
\textit{[Gold labels]} \\
Explanations: \\
\textit{[Gold Explanations]} \\

\end{minipage}
}
\label{fig:propmt_1shot}
\end{figure*}

\section{Model output examples}

In this section, we provide some specific examples of LLM model output to illustrate their strengths and weaknesses. 

\subsection{Task 1: Misinformation span detection \label{app:task1modeloutput}}

In general, many misinformation spans identified by the models tend to be longer than the gold standard ones. For example, while some gold standard spans consist of single nouns, verbs and adjectives, the corresponding model-identified spans often consist of complete noun phrases or verbs and their objects, even if some words are shared between the real and fake articles.  For example, given the sentences \textit{\textbf{Real:} California's clean-air agency voted on Friday to push ahead with stricter emissions standards for cars and trucks} and \textit{\textbf{Fake:} California's clean-air agency voted on Friday to reevaluate their stricter emissions standards for cars and trucks}, the gold standard identifies the fake span \textit{reevaluate} as being inconsistent with \textit{push ahead} in the real article. However, several LLMs identified the misinformation span as \textit{reevaluate their stricter emissions standards}.

\subsection{Task 2: Misinformation classification \label{app:task2modeloutput}}

The output of certain reasoning-oriented models suggests that they may be ``overthinking''' the label assignment process. As an example, the real news headline \textit{Toshiba's Westinghouse files for US bankruptcy} shows that Westinghouse is is trouble, while the corresponding fake headline is \textit{Toshiba's Westinghouse creating thriving job market for US citizens}, which implies that Westinghouse is doing very well. One of the reasoning-oriented models, i.e., Claude Sonnet 4.5, correctly labels the span \textit{creating thriving job market for US citizens} as \textit{Inconsistency}, reasoning that \textit{Real news shows bankruptcy and losses, opposite of thriving}. The comparable sentence structures of the real and fake headlines presumably also make the inconsistency straightforward for the no-think models to spot. In the zero-shot setting, however, another reasoning-oriented model, DeepSeek-Reasoner, classifies this as "No Mention".  While the output of this model initially suggests that it understands the inconsistency: \textit{No mention of creating a thriving job market. In fact, it's in crisis}, it ultimately decides that \textit{It's "No mention" because the real news doesn't talk about job market creation or thriving jobs}. It thus appears that focus on "fine details", i,e., that \textit{bankruptcy} and \textit{job market} are different types of entities, clouds the model's ability to see the inconsistency.    

\subsection{Task 3: Misinformation explanation \label{app:task3modeloutput}}

A ``good'' inconsistency explanation should not only identify the corresponding spans in real and fake stories that lead to the inconsistency, but should also make it clear why this constitutes an inconsistency, especially if this is not immediately clear from the spans themselves. While the output of some models  is almost exclusively limited to identifying the text span in the real article that is inconsistent with the misinformation span, Claude Sonnet 4.5 output often expands upon this by explaining \textit{why} there is an inconsistency, e.g., \textit{Real content says "Brands are also keen to see a true rival emerge to challenge Facebook and Google", which describes brands' desire for a challenge, while fake content says "Facebook and google will be challenged", which presents it as a certainty}. This may help to explain the high similarity score attained by Claude Sonnet 4.5 for Task 3.

\begin{table*}[htb]
\resizebox{\textwidth}{!}{
\begin{tabular}{lllllllllllllllll}
\hline
                  & \multicolumn{8}{c}{Exact}                                                                                                                                                                                                  & \multicolumn{8}{c}{Relax}                                                                                                                                                                                                  \\
Models            & \multicolumn{1}{c}{biz} & \multicolumn{1}{c}{edu} & \multicolumn{1}{c}{entmt} & \multicolumn{1}{c}{polit} & \multicolumn{1}{c}{sports} & \multicolumn{1}{c}{tech} & \multicolumn{1}{c}{cele} & \multicolumn{1}{c}{overall} & \multicolumn{1}{c}{biz} & \multicolumn{1}{c}{edu} & \multicolumn{1}{c}{entmt} & \multicolumn{1}{c}{polit} & \multicolumn{1}{c}{sports} & \multicolumn{1}{c}{tech} & \multicolumn{1}{c}{cele} & \multicolumn{1}{c}{overall} \\ \hline
                  & \multicolumn{16}{c}{\textit{\textbf{Zero-shot}}}                                                                                                                                                                                                                                                                                                                                                                                                        \\ \hline
gpt-5             & 0.306                   & 0.291                   & 0.220                     & \textbf{0.382}            & 0.291                      & 0.250                    & 0.384                    & 0.267                       & 0.433                   & 0.323                   & 0.446                     & 0.391                     & 0.443                      & 0.458                    & 0.308                    & 0.371                       \\
DeepSeek-Reasoner & \textbf{0.458}          & \textbf{0.308}          & \textbf{0.328}            & 0.367                     & \textbf{0.395}             & 0.308                    & 0.350                    & \textbf{0.302}              & \textbf{0.586}          & 0.463                   & 0.541                     & \textbf{0.664}            & 0.458                      & 0.350                    & 0.511                    & \textbf{0.429}              \\
claude-sonnet-4.5 & 0.387                   & 0.292                   & 0.311                     & 0.331                     & 0.314                      & 0.170                    & 0.204                    & 0.300                       & 0.511                   & 0.462                   & 0.481                     & 0.291                     & 0.264                      & \textbf{0.576}           & 0.511                    & \textbf{0.429}              \\
gemini-2.5-flash  & 0.311                   & 0.242                   & 0.322                     & 0.314                     & 0.258                      & 0.323                    & 0.328                    & 0.287                       & 0.423                   & \textbf{0.513}          & 0.405                     & 0.462                     & \textbf{0.688}             & 0.446                    & 0.328                    & 0.420                       \\
Qwen3-8b-R        & 0.369                   & 0.242                   & 0.264                     & 0.367                     & 0.308                      & 0.250                    & 0.424                    & 0.279                       & 0.511                   & 0.463                   & 0.511                     & 0.519                     & 0.476                      & 0.494                    & 0.381                    & 0.388                       \\
Qwen3-14b-R       & 0.291                   & 0.188                   & 0.188                     & 0.338                     & 0.367                      & \textbf{0.404}           & 0.331                    & 0.271                       & 0.387                   & 0.264                   & \textbf{0.546}            & 0.538                     & 0.414                      & 0.511                    & 0.371                    & 0.375                       \\
Qwen3-32b-R       & 0.331                   & 0.224                   & 0.304                     & 0.338                     & 0.264                      & 0.384                    & 0.371                    & 0.271                       & 0.405                   & 0.463                   & 0.463                     & 0.481                     & 0.441                      & 0.529                    & 0.371                    & 0.379                       \\
gpt-4.1           & 0.369                   & 0.242                   & 0.292                     & 0.291                     & 0.349                      & 0.367                    & \textbf{0.436}           & 0.291                       & 0.519                   & 0.395                   & 0.528                     & 0.450                     & 0.598                      & 0.414                    & 0.511                    & 0.426                       \\
DeepSeek-Chat     & 0.350                   & 0.242                   & 0.250                     & 0.367                     & 0.291                      & 0.323                    & 0.369                    & 0.286                       & 0.476                   & 0.409                   & 0.478                     & 0.576                     & 0.414                      & 0.476                    & 0.381                    & 0.398                       \\
Qwen3-8b          & 0.188                   & 0.133                   & 0.188                     & 0.264                     & 0.235                      & 0.235                    & 0.264                    & 0.211                       & 0.188                   & 0.250                   & 0.350                     & 0.316                     & 0.350                      & 0.250                    & 0.546                    & 0.285                       \\
Qwen3-14b         & 0.311                   & 0.259                   & 0.235                     & 0.308                     & 0.264                      & 0.308                    & 0.292                    & 0.270                       & 0.423                   & 0.395                   & 0.381                     & 0.409                     & 0.365                      & 0.548                    & 0.371                    & 0.381                       \\
Qwen3-32b         & 0.188                   & 0.224                   & 0.250                     & 0.308                     & 0.235                      & 0.235                    & 0.264                    & 0.251                       & 0.264                   & 0.367                   & 0.436                     & 0.350                     & 0.422                      & 0.511                    & 0.339                    & 0.356                       \\
Qwen2.5-72b       & 0.291                   & 0.259                   & 0.278                     & 0.367                     & 0.278                      & 0.235                    & 0.381                    & 0.268                       & 0.405                   & 0.463                   & 0.463                     & 0.339                     & 0.367                      & 0.511                    & 0.381                    & 0.377                       \\
LLama8b           & 0.224                   & 0.152                   & 0.250                     & 0.264                     & 0.264                      & 0.250                    & 0.220                    & 0.212                       & 0.276                   & 0.350                   & 0.339                     & 0.361                     & 0.291                      & 0.204                    & 0.371                    & 0.314                       \\
LLama70b          & 0.387                   & 0.259                   & 0.264                     & 0.338                     & 0.367                      & 0.235                    & 0.381                    & 0.281                       & 0.511                   & 0.436                   & 0.492                     & 0.353                     & 0.533                      & 0.398                    & \textbf{0.567}           & 0.391                       \\ \hline
                  & \multicolumn{16}{c}{\textit{\textbf{1-shot}}}                                                                                                                                                                                                                                                                                                                                                                                                           \\ \hline
gpt-5             & 0.114                   & \textbf{0.470}          & 0.323                     & 0.353                     & 0.308                      & 0.291                    & 0.235                    & 0.269                       & 0.357                   & 0.422                   & 0.462                     & 0.391                     & 0.291                      & 0.369                    & 0.331                    & 0.374                       \\
DeepSeek-Reasoner & \textbf{0.476}          & 0.353                   & 0.304                     & 0.381                     & \textbf{0.381}             & 0.264                    & 0.404                    & \textbf{0.304}              & \textbf{0.607}          & 0.463                   & 0.560                     & \textbf{0.617}            & \textbf{0.519}             & 0.328                    & \textbf{0.494}           & \textbf{0.433}              \\
claude-sonnet-4.5 & 0.259                   & 0.456                   & \textbf{0.331}            & 0.292                     & 0.331                      & 0.264                    & 0.314                    & 0.294                       & 0.567                   & 0.423                   & 0.381                     & 0.462                     & 0.414                      & 0.519                    & 0.476                    & 0.413                       \\
gemini-2.5-flash  & 0.291                   & 0.311                   & 0.308                     & 0.278                     & 0.331                      & 0.278                    & 0.292                    & 0.280                       & 0.491                   & 0.450                   & 0.446                     & 0.450                     & 0.478                      & 0.501                    & 0.353                    & 0.400                       \\
Qwen3-8b-R        & 0.350                   & 0.308                   & 0.291                     & \textbf{0.395}            & 0.338                      & 0.424                    & 0.276                    & 0.284                       & 0.441                   & \textbf{0.478}          & 0.492                     & 0.579                     & 0.367                      & 0.567                    & 0.350                    & 0.398                       \\
Qwen3-14b-R       & 0.331                   & 0.264                   & 0.316                     & 0.338                     & 0.338                      & \textbf{0.462}           & 0.387                    & 0.256                       & 0.423                   & 0.391                   & \textbf{0.585}            & 0.557                     & 0.476                      & 0.476                    & 0.391                    & 0.338                       \\
Qwen3-32b-R       & 0.423                   & 0.264                   & 0.278                     & 0.338                     & 0.353                      & 0.443                    & 0.259                    & 0.269                       & 0.511                   & 0.371                   & 0.565                     & 0.557                     & 0.441                      & 0.529                    & 0.400                    & 0.373                       \\
gpt-4.1           & 0.384                   & 0.258                   & 0.308                     & 0.250                     & 0.278                      & 0.331                    & 0.338                    & 0.289                       & 0.533                   & 0.353                   & 0.350                     & 0.494                     & 0.436                      & \textbf{0.621}           & 0.484                    & 0.419                       \\
DeepSeek-Chat     & 0.350                   & 0.276                   & 0.264                     & 0.353                     & 0.328                      & 0.220                    & \textbf{0.443}           & 0.296                       & 0.494                   & 0.422                   & 0.541                     & 0.519                     & 0.398                      & 0.395                    & 0.478                    & 0.409                       \\
Qwen3-8b          & 0.133                   & 0.123                   & 0.250                     & 0.278                     & 0.204                      & 0.235                    & 0.188                    & 0.216                       & 0.166                   & 0.170                   & 0.381                     & 0.250                     & 0.291                      & 0.364                    & 0.133                    & 0.294                       \\
Qwen3-14b         & 0.387                   & 0.220                   & 0.220                     & 0.323                     & 0.291                      & 0.404                    & 0.338                    & 0.273                       & 0.494                   & 0.339                   & 0.409                     & 0.462                     & 0.478                      & 0.478                    & 0.304                    & 0.385                       \\
Qwen3-32b         & 0.235                   & 0.242                   & 0.278                     & 0.264                     & 0.250                      & 0.264                    & 0.235                    & 0.262                       & 0.316                   & 0.395                   & 0.339                     & 0.339                     & 0.367                      & 0.511                    & 0.361                    & 0.364                       \\
Qwen2.5-72b       & 0.350                   & 0.259                   & 0.278                     & 0.291                     & 0.278                      & 0.308                    & 0.276                    & 0.271                       & 0.476                   & 0.422                   & 0.381                     & 0.381                     & 0.414                      & 0.462                    & 0.361                    & 0.384                       \\
LLama8b           & 0.220                   & 0.220                   & 0.278                     & 0.304                     & 0.292                      & 0.308                    & 0.404                    & 0.223                       & 0.316                   & 0.381                   & 0.436                     & 0.409                     & 0.481                      & 0.339                    & 0.384                    & 0.310                       \\
LLama70b          & 0.405                   & 0.276                   & 0.278                     & 0.291                     & 0.323                      & 0.292                    & 0.220                    & 0.243                       & 0.511                   & 0.436                   & 0.450                     & 0.367                     & 0.395                      & 0.338                    & 0.316                    & 0.317                       \\ \hline
\end{tabular}
}
\caption{Results on seven domains in Task 1. \label{tab:task1_domsins}}
\end{table*}

\begin{table*}[]
\resizebox{\textwidth}{!}{
\begin{tabular}{lllllllllllllllll}
\hline
                  & \multicolumn{2}{c}{biz}                          & \multicolumn{2}{c}{edu}                          & \multicolumn{2}{c}{entmt}                        & \multicolumn{2}{c}{polit}                        & \multicolumn{2}{c}{sports}                       & \multicolumn{2}{c}{tech}                         & \multicolumn{2}{c}{cele}                                        & \multicolumn{2}{c}{overall}                      \\
Models            & \multicolumn{1}{c}{Acc} & \multicolumn{1}{c}{F1} & \multicolumn{1}{c}{Acc} & \multicolumn{1}{c}{F1} & \multicolumn{1}{c}{Acc} & \multicolumn{1}{c}{F1} & \multicolumn{1}{c}{Acc} & \multicolumn{1}{c}{F1} & \multicolumn{1}{c}{Acc} & \multicolumn{1}{c}{F1} & \multicolumn{1}{c}{Acc} & \multicolumn{1}{c}{F1} & \multicolumn{1}{c}{Acc} & \multicolumn{1}{c}{F1}                & \multicolumn{1}{c}{Acc} & \multicolumn{1}{c}{F1} \\ \hline
                  & \multicolumn{16}{c}{\textit{\textbf{Zero-shot}}}                                                                                                                                                                                                                                                                                                                                                                                     \\ \hline
gpt-5             & 0.487                   & 0.277                  & 0.487                   & 0.320                  & 0.846                   & 0.798                  & 0.564                   & 0.438                  & 0.513                   & 0.498                  & 0.769                   & 0.746                  & 0.795                   & \textbf{0.750}                        & 0.547                   & 0.366                  \\
DeepSeek-Reasoner & 0.538                   & 0.458                  & 0.744                   & \textbf{0.672}         & 0.821                   & 0.717                  & 0.641                   & 0.477                  & 0.564                   & 0.438                  & 0.692                   & 0.619                  & 0.795                   & 0.687                                 & 0.593                   & 0.381                  \\
claude-sonnet-4.5 & 0.718                   & 0.491                  & 0.718                   & 0.539                  & \textbf{0.872}          & \textbf{0.868}         & \textbf{0.744}          & \textbf{0.586}         & 0.564                   & \textbf{0.553}         & \textbf{0.846}          & \textbf{0.794}         & \textbf{0.821}          & 0.607                                 & \textbf{0.644}          & \textbf{0.455}         \\
gemini-2.5-flash  & 0.513                   & 0.426                  & 0.487                   & 0.278                  & 0.718                   & 0.606                  & 0.615                   & 0.503                  & 0.385                   & 0.373                  & 0.590                   & 0.585                  & 0.641                   & 0.381                                 & 0.462                   & 0.307                  \\
Qwen3-8b-R        & 0.538                   & 0.439                  & 0.692                   & 0.373                  & 0.692                   & 0.351                  & 0.692                   & 0.501                  & 0.231                   & 0.164                  & 0.590                   & 0.526                  & 0.744                   & 0.505                                 & 0.510                   & 0.320                  \\
Qwen3-14b-R       & 0.564                   & 0.410                  & 0.744                   & 0.452                  & 0.718                   & 0.445                  & 0.667                   & 0.407                  & 0.641                   & 0.478                  & 0.564                   & 0.590                  & 0.641                   & 0.324                                 & 0.543                   & 0.351                  \\
Qwen3-32b-R       & 0.615                   & 0.429                  & 0.692                   & 0.424                  & 0.692                   & 0.543                  & 0.692                   & 0.518                  & \textbf{0.667}          & 0.501                  & 0.487                   & 0.411                  & 0.667                   & 0.455                                 & 0.528                   & 0.345                  \\
gpt-4.1           & 0.692                   & \textbf{0.541}         & 0.564                   & 0.331                  & \textbf{0.872}          & 0.839                  & 0.692                   & 0.514                  & 0.564                   & 0.465                  & 0.795                   & 0.701                  & 0.692                   & 0.352                                 & 0.593                   & 0.389                  \\
DeepSeek-Chat     & 0.641                   & 0.370                  & \textbf{0.795}          & 0.425                  & 0.744                   & 0.625                  & 0.692                   & 0.552                  & 0.590                   & 0.532                  & 0.692                   & 0.636                  & 0.744                   & 0.373                                 & 0.628                   & 0.436                  \\
Qwen3-8b          & 0.744                   & 0.411                  & 0.692                   & 0.505                  & 0.692                   & 0.405                  & 0.667                   & 0.493                  & 0.564                   & 0.414                  & 0.590                   & 0.493                  & 0.692                   & 0.349                                 & 0.598                   & 0.407                  \\
Qwen3-14b         & \textbf{0.795}          & 0.437                  & 0.641                   & 0.375                  & 0.769                   & 0.536                  & 0.538                   & 0.385                  & 0.487                   & 0.328                  & 0.641                   & 0.561                  & 0.667                   & 0.275                                 & 0.520                   & 0.345                  \\
Qwen3-32b         & 0.564                   & 0.414                  & 0.615                   & 0.388                  & 0.769                   & 0.553                  & 0.641                   & 0.479                  & 0.667                   & 0.469                  & 0.538                   & 0.313                  & 0.692                   & 0.374                                 & 0.548                   & 0.351                  \\
Qwen2.5-72b       & 0.615                   & 0.485                  & 0.692                   & 0.452                  & 0.667                   & 0.583                  & 0.667                   & 0.494                  & 0.487                   & 0.421                  & 0.692                   & 0.583                  & 0.769                   & 0.471                                 & 0.595                   & 0.375                  \\
LLama8b           & 0.564                   & 0.362                  & 0.718                   & 0.490                  & 0.667                   & 0.505                  & 0.590                   & 0.398                  & 0.436                   & 0.360                  & 0.538                   & 0.434                  & 0.615                   & 0.200                                 & 0.549                   & 0.377                  \\
LLama70b          & 0.718                   & 0.489                  & 0.615                   & 0.419                  & 0.692                   & 0.570                  & 0.641                   & 0.482                  & 0.564                   & 0.471                  & 0.615                   & 0.530                  & 0.744                   & 0.561                                 & 0.562                   & 0.357                  \\ \hline
\textbf{}         & \multicolumn{16}{c}{\textit{\textbf{1-shot}}}                                                                                                                                                                                                                                                                                                                                                                                        \\ \hline
gpt-5             & 0.564                   & 0.331                  & 0.615                   & 0.529                  & \textbf{0.897}          & \textbf{0.875}         & 0.667                   & 0.506                  & 0.615                   & 0.376                  & 0.718                   & 0.578                  & \textbf{0.744}          & 0.508                                 & 0.599                   & 0.410                  \\
DeepSeek-Reasoner & 0.564                   & 0.408                  & 0.692                   & 0.461                  & 0.821                   & 0.717                  & 0.641                   & 0.491                  & \textbf{0.744}          & 0.546                  & 0.615                   & 0.504                  & 0.718                   & 0.428                                 & 0.629                   & 0.434                  \\
claude-sonnet-4.5 & 0.667                   & \textbf{0.734}         & 0.769                   & \textbf{0.534}         & 0.846                   & 0.830                  & \textbf{0.744}          & \textbf{0.586}         & 0.513                   & 0.471                  & 0.718                   & 0.696                  & 0.718                   &  \textbf{0.572} & \textbf{0.649}          & \textbf{0.480}         \\
gemini-2.5-flash  & 0.487                   & 0.453                  & 0.564                   & 0.358                  & 0.692                   & 0.527                  & 0.564                   & 0.555                  & 0.667                   & 0.488                  & 0.538                   & 0.496                  & 0.615                   & 0.378                                 & 0.493                   & 0.316                  \\
Qwen3-8b-R        & 0.385                   & 0.340                  & 0.538                   & 0.153                  & 0.692                   & 0.436                  & 0.692                   & 0.501                  & 0.513                   & 0.333                  & 0.487                   & 0.441                  & 0.615                   & 0.430                                 & 0.486                   & 0.312                  \\
Qwen3-14b-R       & 0.615                   & 0.429                  & 0.641                   & 0.404                  & 0.667                   & 0.454                  & 0.564                   & 0.396                  & 0.462                   & 0.314                  & 0.641                   & 0.531                  & 0.692                   & 0.459                                 & 0.528                   & 0.334                  \\
Qwen3-32b-R       & 0.615                   & 0.470                  & 0.641                   & 0.383                  & 0.718                   & 0.608                  & 0.744                   & 0.571                  & 0.615                   & 0.452                  & 0.462                   & 0.321                  & 0.718                   & 0.496                                 & 0.529                   & 0.343                  \\
gpt-4.1           & 0.692                   & 0.604                  & 0.615                   & 0.478                  & 0.872                   & 0.830                  & 0.538                   & 0.409                  & 0.667                   & 0.484                  & \textbf{0.795}          & \textbf{0.740}         & 0.692                   & 0.358                                 & 0.617                   & 0.412                  \\
DeepSeek-Chat     & 0.641                   & 0.494                  & \textbf{0.769}          & 0.524                  & 0.769                   & 0.651                  & 0.641                   & 0.491                  & 0.641                   & \textbf{0.621}         & 0.718                   & 0.699                  & 0.718                   & 0.557                                 & 0.645                   & 0.474                  \\
Qwen3-8b          & \textbf{0.821}          & 0.531                  & 0.718                   & 0.502                  & 0.590                   & 0.249                  & 0.564                   & 0.338                  & 0.564                   & 0.335                  & 0.487                   & 0.332                  & 0.692                   & 0.205                                 & 0.544                   & 0.343                  \\
Qwen3-14b         & 0.718                   & 0.419                  & 0.641                   & 0.392                  & 0.641                   & 0.417                  & 0.667                   & 0.496                  & 0.538                   & 0.380                  & 0.590                   & 0.520                  & 0.641                   & 0.367                                 & 0.440                   & 0.322                  \\
Qwen3-32b         & 0.795                   & 0.472                  & 0.641                   & 0.360                  & 0.718                   & 0.521                  & 0.667                   & 0.497                  & 0.692                   & 0.477                  & 0.538                   & 0.310                  & 0.692                   & 0.354                                 & 0.582                   & 0.350                  \\
Qwen2.5-72b       & 0.821                   & 0.473                  & 0.641                   & 0.291                  & 0.692                   & 0.452                  & 0.718                   & 0.537                  & 0.590                   & 0.417                  & 0.590                   & 0.498                  & 0.667                   & 0.271                                 & 0.598                   & 0.404                  \\
LLama8b           & 0.282                   & 0.111                  & 0.282                   & 0.111                  & 0.436                   & 0.178                  & 0.231                   & 0.225                  & 0.333                   & 0.189                  & 0.282                   & 0.132                  & 0.256                   & 0.133                                 & 0.323                   & 0.279                  \\
LLama70b          & 0.795                   & 0.523                  & 0.667                   & 0.511                  & 0.718                   & 0.574                  & 0.718                   & 0.579                  & 0.564                   & 0.363                  & 0.692                   & 0.600                  & 0.744                   & 0.537                                 & 0.593                   & 0.396                  \\ \hline
\end{tabular}
}
\caption{Results on seven domains in Task 2. \label{tab:task2_domsins}}
\end{table*}

\begin{table*}[]
\resizebox{\textwidth}{!}{
\begin{tabular}{lllllllllllllllll}
\hline
                  & \multicolumn{2}{c}{biz}                             & \multicolumn{2}{c}{edu}                             & \multicolumn{2}{c}{entmt}                           & \multicolumn{2}{c}{polit}                           & \multicolumn{2}{c}{sports}                          & \multicolumn{2}{c}{tech}                            & \multicolumn{2}{c}{cele}                            & \multicolumn{2}{c}{overall}                         \\
Models            & \multicolumn{1}{c}{Rouge} & \multicolumn{1}{c}{Sim} & \multicolumn{1}{c}{Rouge} & \multicolumn{1}{c}{Sim} & \multicolumn{1}{c}{Rouge} & \multicolumn{1}{c}{Sim} & \multicolumn{1}{c}{Rouge} & \multicolumn{1}{c}{Sim} & \multicolumn{1}{c}{Rouge} & \multicolumn{1}{c}{Sim} & \multicolumn{1}{c}{Rouge} & \multicolumn{1}{c}{Sim} & \multicolumn{1}{c}{Rouge} & \multicolumn{1}{c}{Sim} & \multicolumn{1}{c}{Rouge} & \multicolumn{1}{c}{Sim} \\ \hline
                  & \multicolumn{16}{c}{\textit{\textbf{Zero-shot}}}                                                                                                                                                                                                                                                                                                                                                                                              \\ \hline
gpt-5             & 0.488                     & 0.747                   & \textbf{0.720}            & 0.847                   & 0.671                     & 0.781                   & 0.604                     & 0.652                   & 0.480                     & 0.640                   & 0.680                     & 0.837                   & 0.710                     & 0.811                   & 0.618                     & 0.748                   \\
DeepSeek-Reasoner & 0.487                     & 0.744                   & 0.670                     & 0.830                   & 0.606                     & 0.738                   & 0.588                     & 0.639                   & 0.629                     & 0.766                   & 0.683                     & 0.811                   & 0.694                     & 0.811                   & 0.619                     & 0.751                   \\
claude-sonnet-4.5 & \textbf{0.546}            & \textbf{0.844}          & 0.672                     & 0.838                   & 0.675                     & 0.801                   & \textbf{0.726}            & \textbf{0.823}          & 0.512                     & 0.720                   & 0.729                     & 0.859                   & \textbf{0.731}            & \textbf{0.879}          & 0.638                     & \textbf{0.796}          \\
gemini-2.5-flash  & 0.441                     & 0.709                   & 0.630                     & 0.801                   & 0.627                     & 0.775                   & 0.627                     & 0.687                   & 0.581                     & 0.737                   & 0.612                     & 0.763                   & 0.700                     & 0.816                   & 0.593                     & 0.729                   \\
Qwen3-8b-R        & 0.463                     & 0.653                   & 0.657                     & 0.777                   & 0.552                     & 0.701                   & 0.486                     & 0.537                   & 0.524                     & 0.633                   & 0.576                     & 0.716                   & 0.660                     & 0.805                   & 0.528                     & 0.649                   \\
Qwen3-14b-R       & 0.477                     & 0.698                   & 0.654                     & 0.773                   & 0.593                     & 0.728                   & 0.666                     & 0.701                   & 0.626                     & 0.789                   & 0.651                     & 0.746                   & 0.647                     & 0.761                   & 0.573                     & 0.690                   \\
Qwen3-32b-R       & 0.474                     & 0.692                   & 0.593                     & 0.766                   & 0.448                     & 0.560                   & 0.452                     & 0.521                   & 0.463                     & 0.566                   & 0.518                     & 0.569                   & 0.729                     & 0.808                   & 0.546                     & 0.660                   \\
gpt-4.1           & 0.468                     & 0.721                   & 0.714                     & \textbf{0.852}          & 0.667                     & 0.799                   & 0.714                     & 0.788                   & 0.698                     & 0.844                   & 0.676                     & 0.837                   & 0.695                     & 0.836                   & 0.614                     & 0.768                   \\
DeepSeek-Chat     & 0.478                     & 0.706                   & 0.694                     & 0.832                   & 0.591                     & 0.707                   & 0.581                     & 0.649                   & 0.677                     & 0.801                   & \textbf{0.763}            & 0.863                   & 0.700                     & 0.776                   & \textbf{0.649}            & 0.764                   \\
Qwen3-8b          & 0.462                     & 0.696                   & 0.676                     & 0.841                   & 0.571                     & 0.722                   & 0.664                     & 0.718                   & 0.639                     & 0.808                   & 0.730                     & 0.854                   & 0.661                     & 0.822                   & 0.601                     & 0.748                   \\
Qwen3-14b         & 0.470                     & 0.711                   & 0.607                     & 0.792                   & 0.562                     & 0.708                   & 0.647                     & 0.681                   & 0.669                     & 0.807                   & 0.665                     & 0.751                   & 0.644                     & 0.755                   & 0.603                     & 0.733                   \\
Qwen3-32b         & 0.445                     & 0.710                   & 0.634                     & 0.807                   & 0.635                     & 0.752                   & 0.602                     & 0.683                   & 0.574                     & 0.741                   & 0.607                     & 0.730                   & 0.510                     & 0.620                   & 0.583                     & 0.717                   \\
Qwen2.5-72b       & 0.514                     & 0.774                   & 0.683                     & 0.849                   & \textbf{0.695}            & \textbf{0.833}          & 0.684                     & 0.750                   & \textbf{0.709}            & \textbf{0.846}          & 0.717                     & \textbf{0.874}          & 0.657                     & 0.795                   & 0.642                     & 0.785                   \\
LLama8b           & 0.403                     & 0.706                   & 0.567                     & 0.781                   & 0.492                     & 0.666                   & 0.527                     & 0.658                   & 0.491                     & 0.710                   & 0.570                     & 0.758                   & 0.528                     & 0.746                   & 0.481                     & 0.678                   \\
LLama70b          & 0.450                     & 0.759                   & 0.689                     & 0.808                   & 0.576                     & 0.672                   & 0.584                     & 0.656                   & 0.629                     & 0.772                   & 0.639                     & 0.784                   & 0.640                     & 0.772                   & 0.598                     & 0.740                   \\ \hline
                  & \multicolumn{16}{c}{\textit{\textbf{1-shot}}}                                                                                                                                                                                                                                                                                                                                                                                                 \\ \hline
gpt-5             & 0.460                     & 0.745                   & \textbf{0.720}            & \textbf{0.852}          & \textbf{0.728}            & 0.854                   & 0.659                     & 0.770                   & 0.566                     & 0.788                   & 0.694                     & 0.857                   & 0.711                     & 0.858                   & 0.633                     & 0.789                   \\
DeepSeek-Reasoner & 0.506                     & 0.773                   & 0.682                     & 0.838                   & 0.626                     & 0.761                   & 0.609                     & 0.717                   & 0.671                     & 0.814                   & 0.688                     & 0.819                   & 0.692                     & 0.814                   & 0.634                     & 0.779                   \\
claude-sonnet-4.5 & 0.508                     & 0.777                   & 0.663                     & 0.837                   & 0.727                     & \textbf{0.865}          & \textbf{0.749}            & \textbf{0.852}          & 0.639                     & 0.824                   & 0.680                     & 0.842                   & \textbf{0.760}            & 0.877                   & 0.640                     & \textbf{0.802}          \\
gemini-2.5-flash  & 0.529                     & 0.828                   & 0.640                     & 0.825                   & 0.680                     & 0.847                   & 0.747                     & \textbf{0.852}          & 0.492                     & 0.684                   & 0.575                     & 0.778                   & 0.678                     & \textbf{0.881}          & 0.610                     & 0.766                   \\
Qwen3-8b-R        & \textbf{0.577}            & 0.810                   & 0.676                     & 0.797                   & 0.350                     & 0.485                   & 0.701                     & 0.796                   & 0.539                     & 0.664                   & 0.250                     & 0.324                   & 0.721                     & 0.840                   & 0.499                     & 0.614                   \\
Qwen3-14b-R       & 0.417                     & 0.731                   & 0.610                     & 0.773                   & 0.432                     & 0.546                   & 0.569                     & 0.637                   & 0.624                     & 0.777                   & 0.683                     & 0.779                   & 0.685                     & 0.799                   & 0.561                     & 0.684                   \\
Qwen3-32b-R       & 0.451                     & 0.717                   & 0.589                     & 0.737                   & 0.577                     & 0.707                   & 0.499                     & 0.604                   & 0.297                     & 0.412                   & 0.456                     & 0.550                   & 0.709                     & 0.815                   & 0.533                     & 0.660                   \\
gpt-4.1           & 0.573                     & \textbf{0.845}          & 0.663                     & 0.842                   & 0.727                     & 0.840                   & 0.631                     & 0.772                   & 0.647                     & 0.832                   & 0.678                     & 0.869                   & 0.704                     & 0.860                   & 0.634                     & 0.796                   \\
DeepSeek-Chat     & 0.529                     & 0.747                   & 0.676                     & 0.839                   & 0.677                     & 0.819                   & 0.610                     & 0.707                   & 0.596                     & 0.766                   & 0.718                     & 0.840                   & 0.716                     & 0.844                   & \textbf{0.641}            & 0.776                   \\
Qwen3-8b          & 0.278                     & 0.413                   & 0.375                     & 0.505                   & 0.222                     & 0.321                   & 0.270                     & 0.327                   & \textbf{0.733}            & 0.830                   & 0.227                     & 0.324                   & 0.450                     & 0.559                   & 0.506                     & 0.639                   \\
Qwen3-14b         & 0.562                     & 0.779                   & 0.637                     & 0.832                   & 0.626                     & 0.755                   & 0.684                     & 0.761                   & 0.587                     & 0.780                   & 0.656                     & 0.788                   & 0.649                     & 0.782                   & 0.609                     & 0.746                   \\
Qwen3-32b         & 0.474                     & 0.747                   & 0.590                     & 0.830                   & 0.635                     & 0.768                   & 0.579                     & 0.691                   & 0.544                     & 0.730                   & 0.605                     & 0.771                   & 0.720                     & 0.845                   & 0.598                     & 0.745                   \\
Qwen2.5-72b       & 0.562                     & 0.819                   & 0.654                     & 0.819                   & 0.690                     & 0.829                   & 0.738                     & 0.836                   & 0.640                     & \textbf{0.836}          & \textbf{0.718}            & \textbf{0.876}          & 0.746                     & 0.878                   & 0.638                     & 0.792                   \\
LLama8b           & 0.259                     & 0.456                   & 0.164                     & 0.301                   & 0.171                     & 0.319                   & 0.281                     & 0.431                   & 0.209                     & 0.306                   & 0.198                     & 0.308                   & 0.275                     & 0.457                   & 0.362                     & 0.516                   \\
LLama70b          & 0.511                     & 0.758                   & 0.660                     & 0.831                   & 0.613                     & 0.789                   & 0.600                     & 0.733                   & 0.565                     & 0.757                   & 0.679                     & 0.832                   & 0.644                     & 0.840                   & 0.609                     & 0.769                   \\ \hline
\end{tabular}
}
\caption{Results on seven domains in Task 3. \label{tab:task3_domsins}}
\end{table*}

\end{document}